\documentclass[manuscript,screen,nonacm]{acmart}

\AtBeginDocument{%
  }

\setcopyright{none}
\usepackage{soul}
\usepackage{subcaption} 
\usepackage{array}

\begin{document}

\captionsetup{
  format=plain,
  labelfont=bf,
  textfont=normalfont,
  labelsep=period,
  singlelinecheck=false
}

\title{Exploration of Reinforcement Learning in Generative Recommendation Models}

\author{PengFei Zheng}
\email{zpf824650@mail.ustc.edu.cn}
\author{Wuchao Li}
\email{liwuchao@mail.ustc.edu.cn}
\affiliation{%
  \institution{University of Science and Technology of China}
  \city{Hefei}
  \country{China}}

\author{Shiyao Wang}
\email{450681821@qq.com}
\author{Honghui Bao}
\email{baohonghui@kuaishou.com}
\affiliation{%
  \institution{Kuaishou}
  \city{Beijing}
  \country{China}}

\author{Hongyi Xie}
\email{hyxie2023@mail.ustc.edu.cn}
\affiliation{%
  \institution{University of Science and Technology of China}
  \city{Hefei}
  \country{China}}

\author{Jiaxin Deng}
\email{dengjiaxin03@kuaishou.com}
\author{Xingmei Wang}
\email{wangxingmei@kuaishou.com}
\author{Jinghao Zhang}
\email{jinghao.zhang@cripac.ia.ac.cn}
\affiliation{%
  \institution{Kuaishou}
  \city{Beijing}
  \country{China}}

\author{Yufei Ye}
\email{aboluo2003@mail.ustc.edu.cn}
\affiliation{%
  \institution{University of Science and Technology of China}
  \city{Hefei}
  \country{China}}

\author{Ruiming Tang}
\email{tangruiming@kuaishou.com}
\author{Guorui Zhou}
\email{guoruizhou1993@gmail.com}
\author{Han Li}
\email{lihan08@kuaishou.com}
\affiliation{%
  \institution{Kuaishou}
  \city{Beijing}
  \country{China}}

\author{Kun Gai}
\email{gai.kun@qq.com}
\affiliation{%
  \institution{Unaffiliated}
  \city{Beijing}
  \country{China}}

\author{Defu Lian}
\authornote{Corresponding author: Defu Lian.}
\email{liandefu@ustc.edu.cn}
\affiliation{%
  \institution{University of Science and Technology of China}
  \city{Hefei}
  \country{China}}

% Keep the additional affiliation in the contact-information footnote only.
\makeatletter
\authorsaddresses{%
  \@mkauthorsaddresses\unskip; also affiliated with the State Key Laboratory of Cognitive Intelligence}
\makeatother

\renewcommand{\shortauthors}{Zheng et al.}

\begin{abstract}
Traditional cascaded recommender systems balance accuracy and efficiency over large item spaces, but misaligned stage-wise objectives hinder end-to-end optimization. Generative recommendation unifies these stages in an autoregressive framework, providing a natural interface for reinforcement learning (RL) to optimize the entire generation process. However, directly applying RL often leads to unstable optimization and degraded recommendation quality.

We revisit Group Relative Policy Optimization (GRPO) and other ratio-based objectives through a unified logit-gradient view. We identify three issues: (i) clipped surrogates create dead-gradient regions, under-utilizing samples; (ii) negative-advantage updates can become effectively unbounded as importance ratios grow, allowing a small subset of negatives to dominate optimization; and (iii) ratio-based control lacks sensitivity to absolute policy confidence, so updates on high-confidence cases may fail to vanish.

Motivated by this diagnosis, we propose \emph{Gradient-Bounded Policy Optimization} (GBPO), a simple objective with stop-gradient normalization that is simultaneously (a) fully gradient-active, (b) sign-symmetrically bounded for both positive and negative updates, and (c) explicitly confidence-weighted under current policy probabilities. This ensures that updates gracefully vanish for $\pi\!\to\!1$ positives and $\pi\!\to\!0$ negatives. Across the evaluated settings, GBPO achieves competitive recommendation performance with more stable training than competing RL methods.
\end{abstract}


%\received{18 April 2026}

\maketitle

\section{Introduction}
Industrial-scale recommendation systems have long relied on multi-stage cascaded architectures---typically comprising retrieval, ranking, and re-ranking---to balance effectiveness and efficiency over massive item spaces~\cite{DBLP:conf/recsys/CovingtonAS16,DBLP:journals/corr/abs-2007-16122}. While successful, these stage-wise pipelines often suffer from misaligned objectives, which complicates end-to-end optimization and limits the system's ability to adapt to dynamic user behaviors~\cite{zhou2025onerectechnicalreport}.

Generative recommendation has recently emerged as a transformative paradigm to address these limitations. By framing recommendation as conditional sequence generation, a single model can unify traditionally separate stages into a cohesive end-to-end framework~\cite{rajput2023recommender,zhou2025onerectechnicalreport,zhou2025onerecv2technicalreport}. Beyond architectural simplification, this formulation also supports reinforcement learning (RL) post-training. Here, we focus on post-training for next-item recommendation, using rewards derived from a frozen teacher model.

Among modern RL methods, Group Relative Policy Optimization (GRPO)~\cite{shao2024deepseekmathpushinglimitsmathematical} has emerged as a widely used ratio-based objective, valued for its ability to optimize policies efficiently without a separate critic model. However, directly applying such ratio-based optimization to generative recommendation often leads to severe training instability and degraded recommendation quality~\cite{DBLP:journals/corr/abs-2510-20150}. To provide a principled explanation for these pathologies, we revisit ratio-based objectives through a unified logit-gradient perspective. We identify three structural issues that are particularly consequential in the recommendation setting: 
(i) \textbf{Dead-gradient regions}: clipped surrogates induce saturation where sampled tokens provide no learning signal, leading to sample under-utilization; 
(ii) \textbf{Unbounded negative updates}: widely used objectives can yield updates that grow indefinitely with the importance ratio on the negative-advantage side, allowing a small subset of negatives to dominate optimization; 
(iii) \textbf{Confidence insensitivity}: ratio-based control lacks sensitivity to absolute policy confidence. Updates on high-confidence cases may fail to vanish when they are driven by ratios to a behavior policy rather than absolute current confidence.

Motivated by this diagnosis, we propose \emph{Gradient-Bounded Policy Optimization} (GBPO), a simple yet effective RL objective tailored for generative recommendation. GBPO introduces a stop-gradient normalization mechanism that achieves three key properties: (a) it is \emph{fully gradient-active}, ensuring all sampled tokens contribute to the learning signal; (b) it is \emph{sign-symmetrically bounded} for both positive and negative updates, preventing unbounded logit gradients; and (c) it is \emph{explicitly confidence-weighted} under the current policy, ensuring updates gracefully vanish as the model gains absolute confidence ($\pi\!\to\!1$ for positives and $\pi\!\to\!0$ for negatives). This design naturally prioritizes hard, informative cases while suppressing updates on overly confident ones.

We evaluate GBPO on multiple public benchmarks. The results show that GBPO is competitive in recommendation accuracy and more stable during training than the evaluated RL objectives.

\paragraph{Contributions.}
Our main contributions are summarized as follows:
\begin{itemize}
    \item \textbf{Unified Gradient View.} We present a unified logit-gradient formulation for ratio-based objectives, enabling a method-agnostic comparison via coefficient maps.
    \item \textbf{Structural Diagnosis.} We identify and analyze three key pathologies of ratio-based objectives in generative recommendation: dead-gradient regions, unbounded negative updates, and lack of absolute confidence sensitivity.
    \item \textbf{GBPO Objective.} We propose Gradient-Bounded Policy Optimization (GBPO), featuring a stop-gradient normalization that yields full sample utilization, sign-symmetric boundedness, and confidence-weighted updates.
    \item \textbf{Empirical Validation.} Experiments across datasets, backbones, and reward settings demonstrate GBPO's competitive recommendation accuracy and training-stability advantage among the evaluated RL objectives.
\end{itemize}

The remainder of the paper is organized as follows. Section~2 reviews related work on generative recommendation, ratio-based RL post-training, and hard-example emphasis. Section~3 formalizes the recommendation-RL setting and introduces the notation used throughout the paper. Section~4 develops the unified logit-gradient view and presents GBPO. Section~5 reports the empirical evaluation, including main results, design choices, and stability analysis. Section~6 presents the ethics statement, Section~7 concludes the paper, and Section~8 discusses limitations.

\begin{figure}[htbp]
    \centering
    \includegraphics[width=0.9\linewidth]{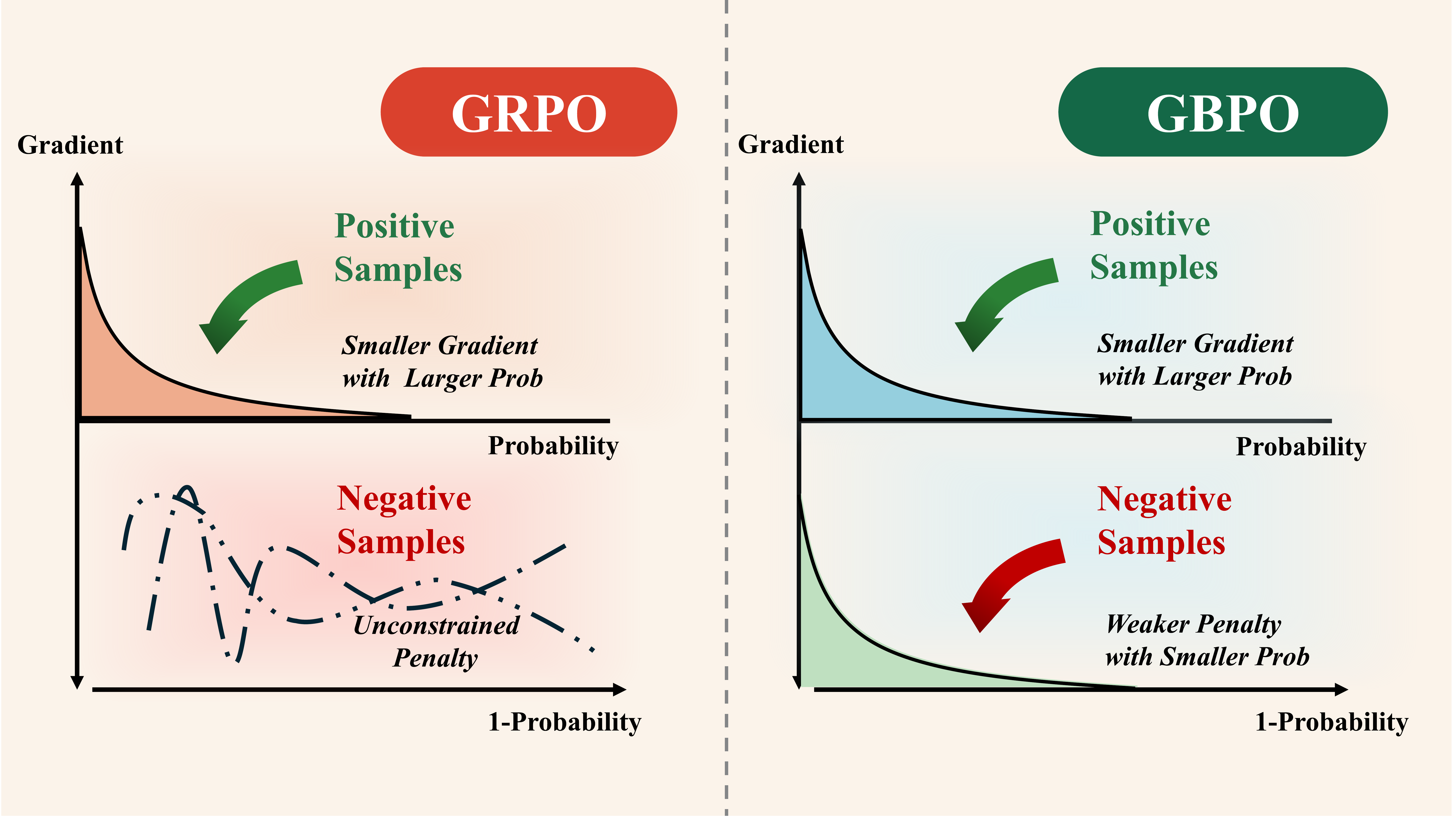}
    \Description{Comparison of Gradient Behavior Between GRPO and GBPO.}
    \caption{
    \textbf{Comparison of Gradient Behavior Between GRPO and GBPO.}
    \textit{Left:} GRPO enforces bounded, confidence-sensitive updates for positive samples but leaves negative-sample gradients unconstrained.
    \textit{Right:} GBPO guarantees bounded gradients for both update signs, with magnitudes that vanish as absolute confidence increases. This design naturally prioritizes hard, informative examples while preventing unbounded logit gradients.
    }
    \label{fig:total_view}
\end{figure}

\section{Related Work}

\subsection{Traditional Recommendation Models}
Industrial-scale recommender systems commonly adopt multi-stage pipelines
(e.g., retrieval and ranking, optionally followed by re-ranking)
to balance effectiveness and serving efficiency~\cite{DBLP:conf/recsys/CovingtonAS16}.
For retrieval, two-tower models learn separate user/item encoders to enable efficient approximate search under dot-product similarity~\cite{DBLP:conf/cikm/HuangHGDAH13,DBLP:conf/recsys/CovingtonAS16}. 
Parallel to this, sequential recommenders (e.g., GRU4Rec, SASRec, and BERT4Rec) have been widely studied for next-item prediction, capturing evolving user interests via recurrent or transformer-based architectures~\cite{hidasi2016sessionbasedrecommendationsrecurrentneural,DBLP:conf/icdm/KangM18,sun2019bert4recsequentialrecommendationbidirectional}.
Ranking then employs more expressive interaction models, such as Wide\&Deep and attention-based architectures like DIN, to capture fine-grained feature interactions and relevance scores~\cite{DBLP:conf/recsys/Cheng0HSCAACCIA16,DBLP:conf/kdd/ZhouZSFZMYJLG18}.
Re-ranking further optimizes list-wise objectives (e.g., diversity) via diversification or subset selection,
including MMR and DPP~\cite{DBLP:conf/sigir/CarbonellG98,kulesza2012determinantal}.
Despite their success, these stage-wise pipelines are typically trained with different objectives and
non-differentiable intermediate interfaces (e.g., top-$K$), which can induce cross-stage mismatch and hinder
end-to-end optimization, motivating holistic paradigms such as generative recommendation~\cite{zhou2025onerectechnicalreport}.

\subsection{Generative Recommendation Models}
Generative recommendation frames recommendation as conditional sequence generation, enabling an autoregressive model to generate items from user histories and providing a natural interface for RL-based post-training.
Recent work improves tokenization and architecture for large discrete catalogs: HSTU~\cite{zhai2024actions} models unified event streams, while TIGER~\cite{rajput2023recommender} generates semantic IDs via residual quantization to enhance semantic sharing and cold-start robustness.
Beyond next-item generation, generative ranking predicts user actions over candidates (e.g., GenRank~\cite{huang2025towards}) and models heterogeneous candidates efficiently (e.g., MTGR~\cite{han2025mtgr}).
More recently, end-to-end systems such as OneRec~\cite{zhou2025onerectechnicalreport,zhou2025onerecv2technicalreport} aim to unify retrieval and ranking through session-level generation. These advances make end-to-end generation practical and, importantly, create a natural entry point for applying post-training and ratio-based RL optimization to recommendation policies.

\subsection{Connections to Generative Retrieval}
\label{subsec:generative_retrieval}
Generative retrieval also predicts identifiers autoregressively: DSI maps queries to document IDs~\cite{tay2022transformer}, NCI learns a neural corpus index~\cite{wang2022neural}, and GenRet learns semantic document identifiers~\cite{sun2023learning}. These methods share identifier design, semantic tokenization, and constrained decoding with generative recommendation, while conditioning on queries rather than user histories and collaborative interactions.

\subsection{Reinforcement Learning for LLM Post-Training}
Reinforcement learning is now a core component in post-training LLMs, with PPO~\cite{schulman2017proximalpolicyoptimizationalgorithms} serving as a foundational algorithm in RLHF~\cite{bai2022training} by constraining policy shifts through ratio-based surrogates.
To reduce reliance on a learned critic and improve scalability, GRPO~\cite{shao2024deepseekmathpushinglimitsmathematical} replaces value-function baselines with group-relative normalization, and GSPO~\cite{zheng2025group} further moves importance weighting to the sequence level to mitigate token-level ratio noise.
Recent variants refine clipping and sampling to improve exploration and prevent collapse: Dual-Clip PPO~\cite{DBLP:conf/aaai/YeLSSZWYYWGCYZS20} caps extreme negative updates to stabilize training under large ratio shifts, while DAPO~\cite{DBLP:conf/nips/YuZZYZYDFLLLLLL25} introduces asymmetric clipping and dynamic sampling strategies. Furthermore, GPPO~\cite{su2025gppo} preserves gradient flow in saturated regions via stop-gradient constructions.
Although these methods achieve strong gains in LLM alignment and reasoning, directly transferring ratio-based objectives to \emph{generative recommendation} is non-trivial, due to imperfect proxy reward signals from teacher models and constrained item-ID token spaces~\cite{DBLP:journals/corr/abs-2510-20150}.

\subsection{Hard Example Mining}
Emphasizing \emph{hard} examples is a widely used principle for improving data efficiency and discriminative learning.
Focal loss~\cite{DBLP:conf/iccv/LinGGHD17} down-weights high-confidence cases and concentrates gradient signal on misclassified instances.
In recommendation and preference optimization, S-DPO~\cite{chen2024softmax} assigns larger weights to negatives that the current model still scores highly, encouraging stronger preference separation. ReRe~\cite{tan2025reinforced} shows that effective hard-negative learning benefits from competitive candidate generation (e.g., constrained beam-style sampling) and difficulty-aware supervision (e.g., ranking-aware rewards), rather than treating all negatives equally.
Relatedly, recent work couples difficulty mining with RL by using a curated hard set for RL exploration, refreshing it across SFT/RL alternation, and replaying solved cases in SFT to consolidate gains~\cite{DBLP:journals/corr/abs-2510-16925}.
Similar motivations also appear in self-supervised recommendation, where informative pair construction and hard negatives are key to representation learning~\cite{DBLP:conf/sigir/WuWF0CLX21}.

Building on this principle, we propose an objective that is explicitly confidence-weighted via its inherent gradient structure. This approach improves stability by naturally emphasizing hard cases while suppressing updates on high-confidence samples without requiring auxiliary heuristic weights.

\section{Preliminaries}
\label{sec:prelim}

\paragraph{Task setting: generative recommendation as an autoregressive policy.}
We study reinforcement learning for \emph{generative recommendation} under an autoregressive policy.
Given a user context $x\sim\mathcal{D}$ (e.g., an interaction history), a policy $\pi_\theta$ parameterized by $\theta$
generates a \emph{token sequence} $o=(o_1,\dots,o_{|o|})$ with likelihood
\begin{equation}
\pi_\theta(o\mid x)=\prod_{t=1}^{|o|}\pi_\theta(o_t\mid x,o_{<t}).
\end{equation}
The generated sequence $o$ represents a recommended item and is evaluated by a bounded scalar reward
$R(x,o)$, which measures the quality of the generated sequence under context $x$.
In our setting, $R$ is provided by an external evaluator such as a ranking/scoring model.

\paragraph{Semantic IDs for items.}
Following~\cite{rajput2023recommender}, we represent each item by a fixed-length \emph{semantic ID (SID)} sequence.
Concretely, each item is mapped to $o=(o_1,\dots,o_L)$ where $L$ is a small constant.
Therefore, recommendation reduces to generating a short token sequence, enabling token-level policy optimization for discrete item identities. 

\paragraph{Constrained decoding.}
SID generation is typically constrained. Let $V$ denote the token vocabulary. At position $t$, only a subset of tokens is valid; we denote the valid set by $\mathcal{V}_t(x,o_{<t})\subseteq V$
and the illegal set by $\mathcal{I}_t(x,o_{<t})=V\setminus \mathcal{V}_t(x,o_{<t})$.
During decoding, we enforce validity by masking tokens in $\mathcal{I}_t$.
We use this notation throughout to describe constrained generation in our experiments.

\paragraph{Reference policy and probability ratios.}
We denote by $\pi_{\mathrm{old}}$ the \emph{reference policy} used for candidate generation,
typically a lagged copy of $\pi_\theta$.
For any generated token $o_t$, we define the token-level probability ratio
\begin{equation}
\label{eq:ratio_prelim}
r_t(\theta)\triangleq
\frac{\pi_\theta(o_t\mid x,o_{<t})}{\pi_{\mathrm{old}}(o_t\mid x,o_{<t})}.
\end{equation}

\paragraph{Ratio-based surrogates.}
Many ratio-based objectives optimize a clipped ratio surrogate using
$r_t(\theta)=\pi_\theta/\pi_{\mathrm{old}}$ to limit policy shifts; we will instantiate concrete objectives in Section~\ref{subsec:objectives}.

\paragraph{GRPO: group-relative advantages.}
In its standard formulation, Group Relative Policy Optimization (GRPO) removes the value function by estimating relative advantages within a sampled group.
For each context $x$, it samples $G$ sequences $\{o_i\}_{i=1}^{G}\sim\pi_{\mathrm{old}}(\cdot\mid x)$ and computes a standardized group-relative advantage
\begin{equation}
\label{eq:grpo_adv_prelim}
\hat A_i
=
\frac{R(x,o_i)-\operatorname{mean}(\{R(x,o_k)\}_{k=1}^{G})}
{\operatorname{std}(\{R(x,o_k)\}_{k=1}^{G})},
\end{equation}
assigning $\hat A_i$ to all tokens in $o_i$.

\paragraph{Token-level notation.}
For each context $x$, we generate $G$ candidate sequences $\{o_i\}_{i=1}^G$ using constrained beam search under $\pi_{\mathrm{old}}$,
where $o_i=(o_{i,1},\dots,o_{i,|o_i|})$.
Let $\mathrm{ctx}_{i,t}$ denote the autoregressive context at step $t$ (i.e., $x$ and prefix $o_{i,<t}$).
We write $\pi_{i,t,j}\triangleq \pi_\theta(j\mid \mathrm{ctx}_{i,t})$ with logits $z_{i,t,j}$ and denote the distribution by
$\boldsymbol{\pi}_{i,t}\triangleq \mathrm{softmax}(z_{i,t})$.
We also use $\pi_{i,t}\triangleq \pi_\theta(o_{i,t}\mid \mathrm{ctx}_{i,t})=\pi_{i,t,o_{i,t}}$ for the selected-token probability.
The token-level ratio is $r_{i,t}(\theta)=\pi_{i,t}/\pi_{\mathrm{old},i,t}$ where $\pi_{\mathrm{old},i,t}\triangleq \pi_{\mathrm{old}}(o_{i,t}\mid \mathrm{ctx}_{i,t})$.
We use $\mathrm{sg}(\cdot)$ for stop-gradient.

\paragraph{Advantage conventions.}
We use $\hat A_i$ for trajectory-level advantages shared across tokens (as in GRPO).
Some variants use token-level advantages $\hat A_{i,t}$ (e.g., via shaping/credit assignment),
and sequence-level objectives may use a scalar $A_i^{b}$ defined per sequence with a baseline.
Our analysis treats these advantages as bounded scalars produced by the training pipeline.

\section{Method}
\label{sec:method}

\noindent\textbf{Overview.}
We first present several widely used ratio-based objectives and adopt a unified logit-gradient view to compare them through their \emph{coefficient maps}. This analysis identifies three structural issues: dead-gradient regions, unbounded negative updates, and lack of sensitivity to absolute policy confidence. Finally, we introduce GBPO, which achieves full sample utilization, sign-symmetric gradient bounds, and confidence-aware weighting.

\begin{figure*}[t]
  \centering
  \begin{subfigure}{0.32\textwidth}
    \centering
    \includegraphics[width=\linewidth]{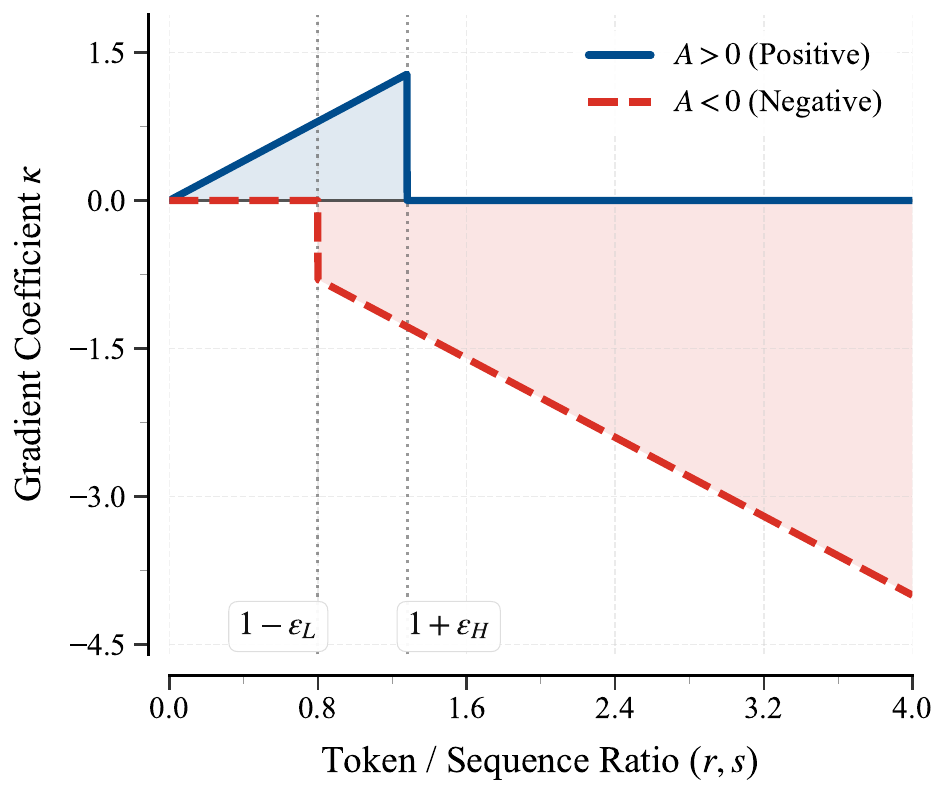}
    \caption{GRPO / GSPO / DAPO (Hard Clip)}
  \end{subfigure}
  \hspace{0.01\textwidth}
  \begin{subfigure}{0.32\textwidth}
    \centering
    \includegraphics[width=\linewidth]{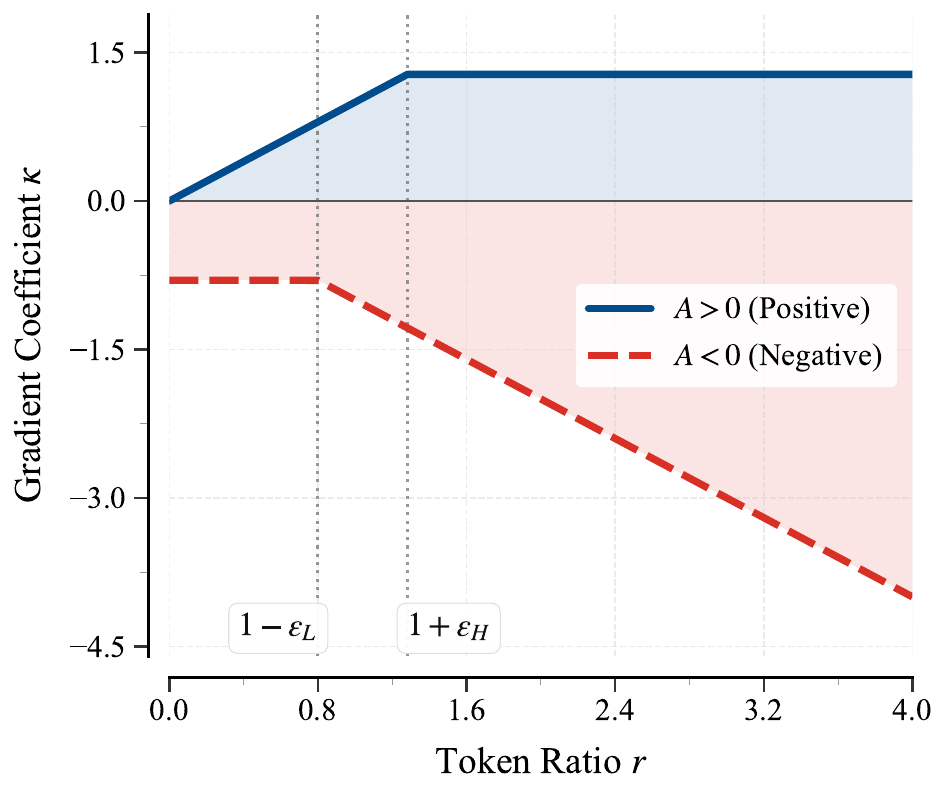}
    \caption{GPPO (Gradient Preserving)}
  \end{subfigure}
  \hspace{0.01\textwidth}
  \begin{subfigure}{0.32\textwidth}
    \centering
    \includegraphics[width=\linewidth]{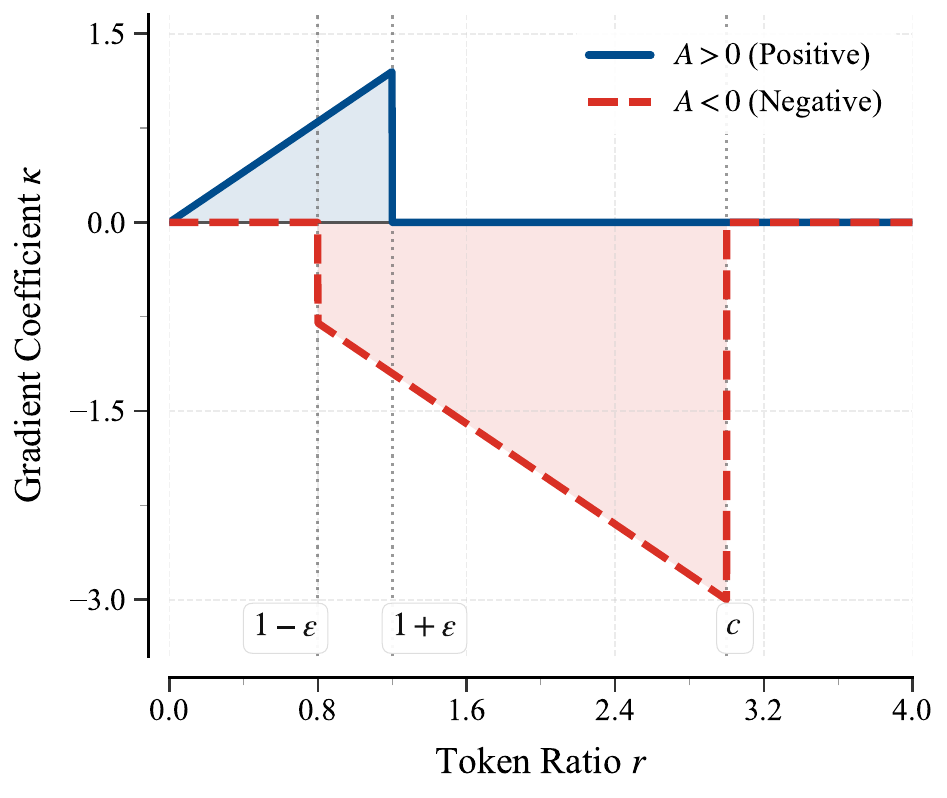}
    \caption{Dual-Clip PPO (Constant Floor)}
  \end{subfigure}
  \caption{Comparison of coefficient maps ($\kappa$) for different policy optimization objectives, showing gradient activity regions across importance ratios and advantages.}
  \Description{Heatmaps showing the gradient activity regions (kappa) for different policy optimization objectives across importance ratios and advantages.}
  \label{fig:all_gradient_regions}
\end{figure*}

\subsection{Representative Policy Optimization Objectives}
\label{subsec:objectives}
We consider several objectives that optimize policy shifts via ratio surrogates.
\paragraph{Clipped objectives.}
GRPO~\cite{shao2024deepseekmathpushinglimitsmathematical} applies sequence-normalized token clipping:
\begin{equation}
\label{eq:loss_grpo}
\mathcal{L}_{\mathrm{GRPO}} = -\frac{1}{G}\sum_{i=1}^{G}\frac{1}{|o_i|}\sum_{t=1}^{|o_i|} \min(r_{i,t}\hat A_i,\, \operatorname{clip}(r_{i,t}, 1-\epsilon, 1+\epsilon)\hat A_i).
\end{equation}
DAPO~\cite{DBLP:conf/nips/YuZZYZYDFLLLLLL25} applies global token normalization ($S \triangleq \sum_{k=1}^G |o_k|$):
\begin{equation}
\label{eq:loss_dapo}
\mathcal{L}_{\mathrm{DAPO}} = -\frac{1}{S}\sum_{i=1}^{G}\sum_{t=1}^{|o_i|} \min(r_{i,t}\hat A_{i,t},\, \operatorname{clip}(r_{i,t}, 1-\epsilon_{\mathrm{low}}, 1+\epsilon_{\mathrm{high}})\hat A_{i,t}).
\end{equation}
GSPO~\cite{zheng2025group} moves weighting to the sequence level via a geometric-mean ratio $\tilde r_i(\theta) = \exp(\frac{1}{|o_i|}\sum_t \log r_{i,t})$ and a sequence-level advantage $A_i^{b}$:
\begin{equation}
\label{eq:loss_gspo}
\mathcal{L}_{\mathrm{GSPO}} = -\frac{1}{G}\sum_{i=1}^{G} \min(\tilde r_i A_i^{b},\, \operatorname{clip}(\tilde r_i, 1-\epsilon, 1+\epsilon)A_i^{b}).
\end{equation}

\paragraph{Refined clipped variants.}
GPPO~\cite{su2025gppo} preserves gradients in saturated regions:
\begin{equation}
\label{eq:loss_gppo}
\mathcal{L}_{\mathrm{GPPO}} = -\frac{1}{S}\sum_{i=1}^{G}\sum_{t=1}^{|o_i|} \min(r_{i,t}\hat A_{i,t},\, \bar r_{i,t}\hat A_{i,t}),
\end{equation}
where $\bar r_{i,t}$ is a stop-gradient clipped ratio that preserves gradient in saturation regions (detailed in Appendix~\ref{app:deriv_gppo}).

Dual-Clip PPO~\cite{DBLP:conf/aaai/YeLSSZWYYWGCYZS20} caps extreme negative updates with sequence-normalized tokens:
\begin{equation}
\label{eq:loss_dual}
\begin{aligned}
\mathcal{L}_{\mathrm{DUAL}} = -\frac{1}{G}\sum_{i=1}^{G}\frac{1}{|o_i|}\sum_{t=1}^{|o_i|} \Bigl[ &\mathbb{I}[\hat A_{i,t}\ge 0] L^{\mathrm{clip}}_{i,t} \\
+ &\mathbb{I}[\hat A_{i,t}<0] \max(L^{\mathrm{clip}}_{i,t},\, c \hat A_{i,t}) \Bigr],
\end{aligned}
\end{equation}
where $L^{\mathrm{clip}}_{i,t} \triangleq \min(r_{i,t} \hat A_{i,t}, \operatorname{clip}(r_{i,t}, 1-\epsilon, 1+\epsilon) \hat A_{i,t})$ is the standard PPO clipped surrogate, and $c > 1$ is a constant capping factor (detailed in Appendix~\ref{app:deriv_dualppo}).

\subsection{A Unified Logit-Gradient View}
\label{subsec:unified_grad}

The objectives considered above differ in clipping form, normalization scope, and optimization granularity (token-level versus sequence-level). As a result, comparing their loss expressions directly can obscure the actual mechanism by which they update the policy. We therefore adopt the selected-logit gradient as a common operational lens: it exposes how each method allocates update mass to sampled tokens and makes dead regions, boundedness, and confidence weighting directly comparable across objectives.

To analyze these objectives systematically, we adopt a unified logit-gradient view. Away from measure-zero non-differentiable points, their gradients share a common \emph{logit-gradient skeleton}:
\begin{equation}
\label{eq:unified_grad_method}
\frac{\partial \mathcal{L}}{\partial z_{i,t,j}} = \alpha_{i,t}\Big(\pi_{i,t,j}-\mathbb{I}[j=o_{i,t}]\Big),
\end{equation}
where $o_{i,t}$ is the sampled token at step $t$ for sequence $i$, and the scalar coefficient $\alpha_{i,t}$ captures the objective's normalization and piecewise behavior. Specifically, $\alpha_{i,t}$ can be decomposed as $\alpha_{i,t} = \eta \cdot \kappa(r, \hat A)$, where $\eta$ is the method-specific normalizer (e.g., $\eta = (G|o_i|)^{-1}$ for GRPO) and $\kappa$ is the \emph{coefficient map} (full forms in Appendix~\ref{app:grad_derivations_full}). Notably, the selected-token logit gradient $| \partial \mathcal{L} / \partial z_{i,t,o_{i,t}} |$ characterizes the update intensity for the entire distribution (proof in Appendix~\ref{app:softmax_identities}), justifying our focus on this term for diagnosis.

%The training behavior is primarily governed by $\kappa$ and its interaction with the selected-token logit gradient (where the softmax Jacobian yields a factor of $1-\pi_{i,t}$): \emph{dead-gradient regions} correspond to $\kappa=0$, \emph{unbounded updates} to $|\kappa(1-\pi_{i,t})| \to \infty$, and \emph{confidence weighting} to the gradient magnitude $|\kappa|(1-\pi_{i,t})$ adapting to the current confidence $\pi_{i,t}$.

\subsection{Diagnostic Criteria and Baseline Analysis}
\label{subsec:diagnosis}
We evaluate these objectives based on three core properties, which are decomposed into five specific criteria in Table~\ref{tab:rl_objective_comparison} to account for both positive and negative advantages (detailed qualitative analysis provided in Appendix~\ref{app:qualitative_analysis}). These properties are derived from the \emph{sampled logit gradient} $\partial \mathcal{L} / \partial z_{i,t,o_{i,t}}$, which incorporates the coefficient map $\kappa$ and the softmax Jacobian term $(1-\pi_{i,t})$. While Figure~\ref{fig:all_gradient_regions} visualizes the behavior of $\kappa$, we define the diagnostic properties as follows:
\begin{itemize}
    \item \textbf{Full sample utilization:} Whether the objective avoids \emph{dead-gradient regions} ($\kappa=0$) for all sampled tokens.
    \item \textbf{Gradient boundedness:} Whether the logit gradient remains bounded independently of extreme ratios (evaluated for both positive and negative advantages in Table~\ref{tab:rl_objective_comparison}).
    \item \textbf{Confidence weighting:} Whether the gradient magnitude vanishes smoothly as a function of absolute confidence $\pi_{i,t}$, without relying on ratio-triggered dead-gradient cutoffs, ensuring updates vanish for high-confidence positives and already-suppressed negatives alike.
\end{itemize}

\paragraph{Analysis of clipped objectives (GRPO, DAPO, GSPO)}
These methods inherit clipped hard truncation, creating significant dead gradient regions once the importance ratio reaches the clip threshold (Figure~\ref{fig:all_gradient_regions} left). In the active regions, their coefficients $\kappa$ remain proportional to importance ratios ($r=r_{i,t}$ for GRPO/DAPO and $r=\tilde r_i$ for GSPO), leading to two critical issues. First, they lack an explicit ratio-independent bound for negative updates; as $r$ grows, the negative gradient can become arbitrarily large. Second, they lack sensitivity to absolute policy confidence. For already-suppressed negatives ($\pi \to 0$), the logit-gradient magnitude behaves as $|\kappa|(1-\pi) \propto r(1-\pi)$. This reveals that updates are driven by \emph{relative} confidence (shifts from the behavior policy), without a reference-independent guarantee of vanishing as the model gains \emph{absolute} confidence, which can destabilize training by over-penalizing tokens that are already improbable.

\paragraph{Analysis of GPPO}
GPPO improves sample utilization by ensuring $\kappa \neq 0$ via a gradient-preserving mechanism (Figure~\ref{fig:all_gradient_regions} middle). Relative to standard clipped surrogates, this eliminates hard zero-gradient truncation and keeps sampled tokens active even after they move beyond the nominal clipping boundary. However, the negative-advantage branch remains ratio-proportional, so GPPO still lacks an explicit ratio-independent bound for negative updates. As a result, it resolves the dead-gradient issue but does not remove the possibility that a small subset of negative samples dominates optimization.

\paragraph{Analysis of Dual-Clip PPO}
Dual-Clip PPO targets the opposite side of the problem: it caps extreme negative updates via a constant floor to mitigate explosion (Figure~\ref{fig:all_gradient_regions} right). This modification improves negative-side stability relative to standard PPO-style clipping, but the floor is constant with respect to the policy and therefore creates new dead-gradient regions once it becomes active. Consequently, Dual-Clip PPO partially addresses negative-update explosion at the cost of reintroducing sample under-utilization, and it still does not achieve consistent absolute confidence-aware weighting for \emph{negative} updates, as summarized in Table~\ref{tab:rl_objective_comparison}.

\begin{table*}[t]
\centering
\small
\caption{\textbf{Qualitative comparison of policy optimization objectives.} Detailed justifications are provided in Appendix~\ref{app:qualitative_analysis}.
``W/ All Samples?'' indicates whether the objective maintains active gradients for all sampled tokens (i.e., avoiding dead-gradient regions).
``Pos/Neg grad bounded'' asks whether the objective prevents arbitrarily large logit gradients.
``Conf.-weighted'' indicates whether the gradient magnitude vanishes smoothly as a function of absolute policy confidence ($\pi \to 1$ for positives, $\pi \to 0$ for negatives). Methods that achieve vanishing solely through confidence-driven decay are marked ``Yes''; those whose vanishing is partly caused by ratio-triggered dead-gradient cutoffs are marked ``Partial''.
For GBPO, both token- and sequence-level variants satisfy the listed qualitative properties.}
\label{tab:rl_objective_comparison}
\setlength{\tabcolsep}{1.5pt}
\renewcommand{\arraystretch}{1.5}
\begin{tabular}{lccccc}
\toprule
\textbf{Method} 
& \textbf{W/ All Samples?} 
& \textbf{Pos-Grad Bounded?} 
& \textbf{Neg-Grad Bounded?} 
& \textbf{Pos Conf.-Weighted?} 
& \textbf{Neg Conf.-Weighted?} \\
\midrule
GRPO  & No  & Yes & No  & Partial & No \\
DAPO  & No  & Yes & No  & Partial & No \\
GSPO  & No  & Yes & No  & Partial & No \\
GPPO  & Yes & Yes & No  & Yes & No \\
Dual-Clip PPO & No & Yes & Yes & Partial & No \\
\midrule
\textbf{GBPO (ours)} & \textbf{Yes} & \textbf{Yes} & \textbf{Yes} & \textbf{Yes} & \textbf{Yes} \\
\bottomrule
\end{tabular}
\end{table*}

\subsection{GBPO: Gradient-Bounded Policy Optimization}
\label{subsec:gbpo}

\paragraph{Motivation from binary cross-entropy.}
Let $p=\sigma(z)$ and $y\in\{0,1\}$. Binary cross-entropy (BCE) is
\begin{equation}
\label{eq:bce_loss}
\ell_{\mathrm{BCE}}(p,y)=-y\log p-(1-y)\log(1-p).
\end{equation}
Using $\partial p/\partial z=p(1-p)$ gives
\begin{equation}
\label{eq:bce_logit_grad}
\frac{\partial\ell_{\mathrm{BCE}}}{\partial z}
=\left(-\frac{y}{p}+\frac{1-y}{1-p}\right)p(1-p)=p-y.
\end{equation}
The positive- and negative-target gradient magnitudes are $1-p$ and $p$, both bounded by one and attenuated for confidently correct predictions. GBPO uses this gradient structure to control policy updates: the advantage sign acts as a pseudo-label and its magnitude supplies the weight.

\paragraph{Token-level GBPO}
The token-level form makes the BCE connection explicit. Let $n_i=|o_i|$, $p_{i,t}=\pi_{i,t}$, and $q_{i,t}=\pi_{\mathrm{old},i,t}$. Define $d_{i,t}=\max(q_{i,t},\mathrm{sg}(p_{i,t}))$ for $\hat A_i\ge0$ and $d_{i,t}=\max(q_{i,t},1-\mathrm{sg}(p_{i,t}))$ for $\hat A_i<0$. The token-mean objective is
\begin{equation}
\label{eq:gbpo_token_obj}
\mathcal L_{\rm token}=-\frac1G\sum_{i=1}^G\frac1{n_i}\sum_t\hat A_i\frac{p_{i,t}}{d_{i,t}}.
\end{equation}
Candidates, advantages, prefixes, reference scores, and denominators are fixed during differentiation; averaging over $B$ contexts adds $1/B$.

\paragraph{Token gradients and the BCE connection.}
For the selected token, writing $p=p_{i,t}$ gives
\begin{equation}
\label{eq:gbpo_token_selected_grad}
g^{\rm token}_{i,t}
=-\frac{\hat A_i}{Gn_i}\frac{p(1-p)}{d_{i,t}}.
\end{equation}
Since $d_{i,t}\ge p$ on the positive branch and $d_{i,t}\ge1-p$ on the negative branch,
\begin{equation}
\label{eq:gbpo_token_bounds}
|g^{\rm token}_{i,t}|\le
\begin{cases}
\dfrac{|\hat A_i|}{Gn_i}(1-p),&\hat A_i\ge0,\\[3pt]
\dfrac{|\hat A_i|}{Gn_i}p,&\hat A_i<0.
\end{cases}
\end{equation}
These are the gradient envelopes of BCE weighted by $|\hat A_i|/(Gn_i)$. With pseudo-label $y_i=\mathbb{I}[\hat A_i>0]$, the selected-logit contribution is exactly $|\hat A_i|\rho_{i,t}(p-y_i)/(Gn_i)$, where $\rho_{i,t}\in(0,1]$ (Appendix~\ref{app:deriv_gbpo}). Thus the reference floor attenuates the BCE-form gradient without reversing its direction or introducing a clipping-induced dead region.

\begin{figure}[htbp]
    \centering
    \includegraphics[width=0.95\linewidth]{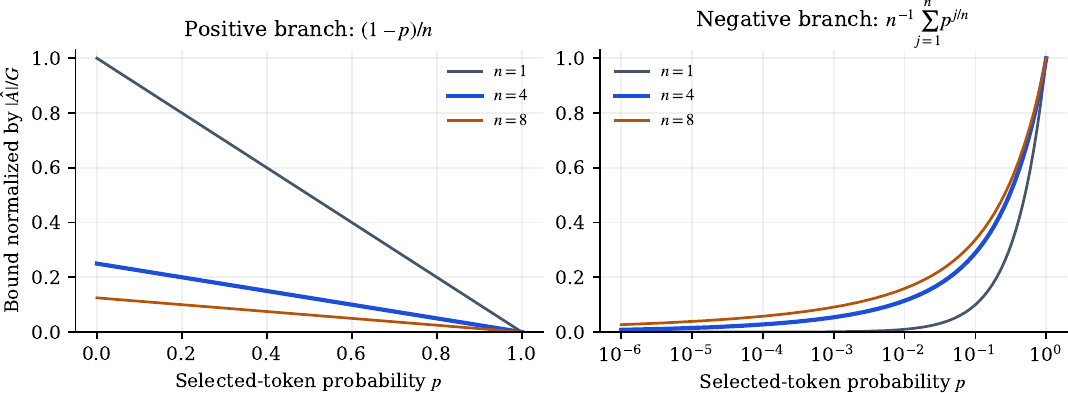}
    \caption{Sequence-level GBPO gradient envelopes, normalized by $|\hat A_i|/G$. For selected-token probability $p$ and identifier length $n$, the positive bound is $(1-p)/n$ and the negative bound is $n^{-1}\sum_{j=1}^{n}p^{j/n}$. Negative-side attenuation weakens as the length increases.}
    \Description{Positive- and negative-advantage selected-logit gradient bounds for identifier lengths 1, 4, and 8.}
    \label{fig:gbpo_confidence_envelopes}
\end{figure}

\paragraph{Sequence-level GBPO used in the main experiments.}
Because rewards are assigned to complete items, we use a shared normalization factor for all tokens in each SID sequence, aligning the optimization granularity with the item-level reward. Our main TIGER experiments apply the same construction to length-normalized sequence scores:
\begin{equation}
\label{eq:gbpo_seq_scores}
P_i=\exp\!\left(\frac{1}{n_i}\sum_t\log p_{i,t}\right),\qquad
Q_i=\exp\!\left(\frac{1}{n_i}\sum_t\log q_{i,t}\right).
\end{equation}
The joint likelihood is $P_i^{n_i}$. Define
\begin{align}
\label{eq:gbpo_norm_pos}
D_i^+&=\max(Q_i,\mathrm{sg}(P_i)), &&\hat A_i\ge0,\\
\label{eq:gbpo_norm_neg}
D_i^-&=\max(Q_i,1-\mathrm{sg}(P_i)), &&\hat A_i<0,
\end{align}
and use the corresponding $D_i$ in
\begin{equation}
\label{eq:gbpo_obj}
\mathcal L_{\mathrm{GBPO}}=-\frac1G\sum_{i=1}^G\hat A_i\frac{P_i}{D_i}.
\end{equation}
Its token-coordinate gradient is
\begin{equation}
\label{eq:gbpo_selected_grad}
\frac{\partial\mathcal L_{\mathrm{GBPO}}}{\partial z_{i,t,j}}
=\frac{\hat A_i}{Gn_i}\frac{P_i}{D_i}
\bigl(\pi_{i,t,j}-\mathbb{I}[j=o_{i,t}]\bigr).
\end{equation}
For the selected-coordinate derivative $g_{i,t}$ and $p=p_{i,t}$, the bounds are
\begin{align}
\label{eq:gbpo_bound_pos}
\hat A_i\ge0:\quad |g_{i,t}|&\le\frac{|\hat A_i|}{Gn_i}(1-p),\\
\label{eq:gbpo_bound_neg}
\hat A_i<0:\quad |g_{i,t}|&\le\frac{|\hat A_i|}{Gn_i}\sum_{j=1}^{n_i}p^{j/n_i}
\le\frac{|\hat A_i|}{G}p^{1/n_i}\le\frac{|\hat A_i|}{G}.
\end{align}
Both signs admit bounds independent of $Q_i$. The negative bound vanishes as $p\to0$ at fixed length, recovering the token bound at $n_i=1$ and weakening for longer identifiers. Table~\ref{tab:rl_objective_comparison} summarizes the qualitative properties; Appendix~\ref{app:deriv_gbpo_seq} gives the full derivation.
Figure~\ref{fig:gbpo_confidence_envelopes} illustrates these confidence-dependent bounds for different SID lengths.

\section{Experiments}
\label{sec:exp}

\subsection{Experimental Setting}
\label{subsec:exp_setting}

\subsubsection{Public Datasets and Evaluation Protocol}
We use three real-world recommendation datasets from the widely used Amazon review benchmark: Beauty, Sports and Toys\footnote{http://jmcauley.ucsd.edu/data/amazon/}.
We follow the standard preprocessing and leave-one-out split protocol used by prior generative recommendation work, where user histories are truncated to a maximum length of $20$ and the last two interactions are reserved for validation and test. Following~\cite{rajput2023recommender}, we report Top-$K$ HitRate and NDCG with $K\in\{5,10\}$.
Unless otherwise stated, we select the checkpoint with the best validation HitRate@10 and report its test performance. This checkpoint-selection protocol is used consistently for both supervised and RL-post-trained models.

\subsubsection{Baselines}
We compare GBPO with representative baselines that cover (i) classic sequential recommenders, (ii) recent generative recommenders, and (iii) non-RL post-training methods.

\textbf{Classic sequential recommenders} include GRU4Rec~\cite{hidasi2016sessionbasedrecommendationsrecurrentneural}, SASRec~\cite{DBLP:conf/icdm/KangM18}, BERT4Rec~\cite{sun2019bert4recsequentialrecommendationbidirectional}, and HGN~\cite{ma2019hierarchicalgatingnetworkssequential}, covering representative RNN-based, transformer-based, and gating-based approaches.

\textbf{Generative recommenders} include TIGER~\cite{rajput2023recommender} and HSTU~\cite{zhai2024actions}, which are competitive state-of-the-art generative models for large discrete catalogs. In our implementation, TIGER generates multi-token SIDs, whereas HSTU directly predicts item IDs, corresponding to $n_i=1$ in the GBPO formulation.

\textbf{Non-RL post-training methods} include listwise knowledge distillation (KD), which matches DIN's distribution over candidates, and S-DPO~\cite{chen2024softmax}, which learns preferences between target items and negative candidates.

\subsubsection{Reward Model and RL Data Generation}
\label{subsubsec:reward_model}
For each dataset, we first train a Deep Interest Network (DIN)~\cite{DBLP:conf/kdd/ZhouZSFZMYJLG18} on the training split and then freeze it during RL. In our implementation, RL training always starts from a supervised checkpoint and uses the frozen DIN only as an external scorer, so the reward model is not updated during policy optimization.
For each context, we generate a set of $G=32$ candidate items. Our training scripts support both rollout sampling and beam generation; the main paper reports results under constrained beam generation from the reference policy $\pi_{\mathrm{old}}$ so that all RL objectives use the same candidate-generation procedure and decoding constraints. Candidate rewards are obtained by ranking DIN scores within each group and assigning binary labels: the top-$K$ candidates receive reward $1$ and the remaining candidates receive reward $0$.
We default to $K=5$ in the main experiments because it provides the best trade-off between reward sparsity and reward noise; the corresponding sensitivity analysis is reported in Section~\ref{subsec:design_choices}. This rank-based reward provides a teacher-guided relative ranking signal while avoiding calibration assumptions on the raw DIN score scale.

\subsubsection{RL Objectives and Implementation Details}
We evaluate GBPO against representative ratio-based objectives, including GRPO~\cite{shao2024deepseekmathpushinglimitsmathematical},
DAPO \cite{DBLP:conf/nips/YuZZYZYDFLLLLLL25}, GSPO~\cite{zheng2025group}, GPPO~\cite{su2025gppo}, and Dual-Clip PPO~\cite{DBLP:conf/aaai/YeLSSZWYYWGCYZS20}.
Unless otherwise stated, all methods within each experiment share the same backbone, decoding constraints, frozen DIN reward model, candidate generation, and optimizer. Our public training scripts use AdamW, $500$ epochs, and eight GPUs for both TIGER and HSTU backbones. The supervised stage uses batch size $1024$, whereas the RL stage uses smaller per-step batches ($128$ for TIGER and $256$ for HSTU) because each batch additionally requires candidate generation and reward evaluation. During evaluation, we decode with a beam size of $256$.
For all compared RL objectives, the reference policy $\pi_{\mathrm{old}}$ is synchronized with the current policy at the beginning of each epoch and remains fixed throughout that epoch.
For each baseline, we follow its original proposed normalization scheme (e.g., sequence-level for GSPO, global token-level for DAPO and GPPO) and additionally tune stability-related hyperparameters to favor baseline performance. In the codebase, GRPO/GSPO use symmetric clipping with $\epsilon=0.2$, DAPO/GPPO use asymmetric clipping $(\epsilon_{\mathrm{low}},\epsilon_{\mathrm{high}})=(0.2,0.28)$, and Dual-Clip PPO uses a clipped PPO surrogate together with a negative floor parameter $c=3$. GBPO is implemented using the sequence-level variant (see Appendix~\ref{app:deriv_gbpo_seq}); Section~\ref{subsec:design_choices} compares this choice against the token-level version.

To further stabilize training and prevent distribution shift, we adopt a joint-objective optimization strategy similar to~\cite{DBLP:journals/corr/abs-2504-05118}. Specifically, the reinforcement learning objective is augmented with an auxiliary supervised fine-tuning (SFT) loss, calculated on the ground-truth recommendation sequences from the training set. The total optimization loss is defined as $\mathcal{L} = \mathcal{L}_{\mathrm{RL}} + \lambda \mathcal{L}_{\mathrm{SFT}}$, where we set the RL loss weight to 1 and the regularization coefficient $\lambda = 0.1$. This formulation serves as a distributional anchor to regularize the RL exploration process, thereby mitigating potential reward-hacking behaviors and enhancing optimization stability.

We define performance collapse (D) as early stopping when validation H@10 falls more than 0.02 below its previous best, and N as a recorded NaN/Inf training loss. In the observed training trajectories, performance did not recover after collapse. Among the 88 collapse events in the repeated default binary-DIN experiments on Amazon and Steam, 75 (85.2\%) involved a final-epoch H@10 drop greater than 0.02 from the preceding epoch. Single-run results use $^\dagger$ (D) and $^\ddagger$ (N); repeated results report D/N as affected-run counts, each out of three. These markers and counts refer to post-training, with SFT entries left blank. Collapsed runs report the best pre-collapse checkpoint.

We use two learning-rate regimes. A high LR ($1\mathrm{e}{-}3$) is used to measure best achievable performance under aggressive optimization, where unstable objectives may collapse. A low LR ($5\mathrm{e}{-}5$) is used for the training-dynamics study so that differences in sample utilization, negative-gradient dominance, and entropy evolution can be observed over a sufficiently long training horizon under a shared budget.

For the repeated RL comparisons, we pair runs by post-training seed (42--44), with the SFT initialization, teacher model, item-ID mapping, and preprocessing fixed within each dataset--backbone setting. We use two-sided paired $t$-tests ($n=3$), applying Holm correction separately to the 20 comparisons (five baselines $\times$ four metrics) within each setting. Appendix~\ref{app:statistical_significance} reports all 140 tests.

Tables~\ref{tab:main_results}--\ref{tab:candidate_size_sensitivity} and~\ref{tab:rl_results} retain the original experimental results. Tables~\ref{tab:non_rl_post_training}--\ref{tab:steam_results} and~\ref{tab:continuous_din_results}--\ref{tab:hstu_three_seed_replication} report experiments conducted on the new server during revision. Table~\ref{tab:statistical_significance} presents statistical tests based on the repeated runs in Tables~\ref{tab:steam_results}, \ref{tab:tiger_three_seed_replication}, and~\ref{tab:hstu_three_seed_replication}.

% =========================================================
\subsection{Main Results}
\label{subsec:main_results}

Table~\ref{tab:main_results} reports the main test results on Amazon datasets.
Across all three datasets, GBPO consistently improves both the TIGER and HSTU supervised backbones under the same architecture and decoding constraints, indicating that the gains can be primarily attributed to more effective post-training optimization.
In particular, compared with TIGER (SFT), TIGER(GBPO) improves HR@5 by +0.69 / +1.96 / +0.80 and HR@10 by +0.71 / +2.21 / +1.23 percentage points on Beauty/Toys/Sports, and improves NDCG@5 by +0.29 / +1.26 / +0.48 and NDCG@10 by +0.30 / +1.35 / +0.62 points. Similarly, HSTU(GBPO) improves over HSTU (SFT) by +0.41 / +0.63 / +0.77 on HR@5, +1.14 / +1.41 / +1.33 on HR@10, +0.03 / +0.10 / +0.41 on NDCG@5, and +0.26 / +0.33 / +0.58 on NDCG@10, showing consistent gains across both backbones.
GBPO also outperforms strong non-generative sequential baselines (e.g., SASRec), suggesting that RL post-training can provide gains beyond architecture/tokenization choices. Notably, TIGER(GBPO) generally outperforms HSTU(GBPO), indicating that the quality of the underlying backbone still influences the final performance ceiling even after RL post-training. To test whether the advantage of GBPO persists beyond a single backbone, we provide a dedicated cross-backbone comparison in Section~\ref{subsec:backbone_generalization}.
An additional pattern is that the gains on Toys and Sports are generally larger than those on Beauty, especially for the generative backbones. A plausible interpretation is that more challenging datasets benefit more from stable RL post-training because informative hard cases are both more common and more consequential. This is consistent with the design of GBPO, which keeps all samples active while allocating stronger learning signal to uncertain, policy-relevant cases rather than over-updating already easy ones.

\begin{table*}[t]
\centering
\small
\caption{\textbf{Main results on Amazon Beauty, Toys, and Sports.}
We report HitRate (H) and NDCG (N) at $\{5, 10\}$ following~\cite{rajput2023recommender}. All numbers are reported at the best validation checkpoint.}
\label{tab:main_results}
\setlength{\tabcolsep}{0.1pt}
\renewcommand{\arraystretch}{1.4}
\begin{tabular}{l *{12}{>{\centering\arraybackslash}p{30pt}}}
\toprule
& \multicolumn{4}{c}{\textbf{Beauty}} 
& \multicolumn{4}{c}{\textbf{Toys}} 
& \multicolumn{4}{c}{\textbf{Sports}} \\
\cmidrule(lr){2-5}\cmidrule(lr){6-9}\cmidrule(lr){10-13}
\textbf{Method}
& \textbf{H@5} & \textbf{H@10} & \textbf{N@5} & \textbf{N@10}
& \textbf{H@5} & \textbf{H@10} & \textbf{N@5} & \textbf{N@10}
& \textbf{H@5} & \textbf{H@10} & \textbf{N@5} & \textbf{N@10} \\
\midrule
GRU4Rec~\cite{hidasi2016sessionbasedrecommendationsrecurrentneural}
& 0.0395 & 0.0584 & 0.0265 & 0.0326
& 0.0330 & 0.0490 & 0.0228 & 0.0279
& 0.0190 & 0.0312 & 0.0122 & 0.0161 \\
SASRec~\cite{DBLP:conf/icdm/KangM18}
& 0.0402 & 0.0607 & 0.0254 & 0.0320
& 0.0448 & 0.0626 & 0.0300 & 0.0358
& 0.0199 & 0.0301 & 0.0106 & 0.0141 \\
HGN~\cite{ma2019hierarchicalgatingnetworkssequential}
& 0.0319 & 0.0536 & 0.0196 & 0.0266
& 0.0326 & 0.0517 & 0.0192 & 0.0254
& 0.0183 & 0.0313 & 0.0109 & 0.0150 \\
BERT4Rec~\cite{sun2019bert4recsequentialrecommendationbidirectional}
& 0.0232 & 0.0396 & 0.0146 & 0.0199
& 0.0215 & 0.0332 & 0.0131 & 0.0168
& 0.0102 & 0.0175 & 0.0065 & 0.0088 \\
\midrule
HSTU~\cite{zhai2024actions} (SFT)
& 0.0457 & 0.0627 & \underline{0.0327} & 0.0382
& 0.0427 & 0.0576 & 0.0322 & 0.0371
& 0.0211 & 0.0317 & 0.0141 & 0.0175 \\
TIGER~\cite{rajput2023recommender} (SFT)
& 0.0454 & 0.0682 & 0.0298 & 0.0371
& 0.0347 & 0.0571 & 0.0224 & 0.0296
& 0.0220 & 0.0366 & 0.0139 & 0.0186 \\
\midrule
\textbf{HSTU(GBPO)}
& \underline{0.0498} & \underline{0.0741} & \textbf{0.0330} & \textbf{0.0408}
& \underline{0.0490} & \underline{0.0717} & \underline{0.0332} & \underline{0.0404}
& \underline{0.0288} & \underline{0.0450} & \underline{0.0182} & \underline{0.0233} \\
\textbf{TIGER(GBPO)}
& \textbf{0.0523} & \textbf{0.0753} & \underline{0.0327} & \underline{0.0401}
& \textbf{0.0543} & \textbf{0.0792} & \textbf{0.0350} & \textbf{0.0431}
& \textbf{0.0300} & \textbf{0.0489} & \textbf{0.0187} & \textbf{0.0248} \\
\bottomrule
\end{tabular}
\end{table*}

% =========================================================
\subsection{Comparison Across RL Objectives}
\label{subsec:rl_objectives_results}

We compare GBPO with representative ratio-based objectives on the TIGER backbone, with results summarized in Table~\ref{tab:rl_results}, under the shared setup in Section~\ref{subsec:exp_setting}.
Under the high-LR regime, we report the best checkpoint by validation HitRate@10 (for collapsed runs, the best pre-collapse checkpoint). Across all three datasets, GBPO consistently outperforms all existing ratio-based RL baselines, including GRPO, DAPO, GSPO, GPPO, and Dual-Clip PPO, on both HitRate and NDCG metrics.
To connect empirical behaviors back to our structural diagnosis, we further analyze detailed training dynamics and stability diagnostics in Section~\ref{subsec:stability_dynamics}.

\subsection{Generalization Across Backbones}
\label{subsec:backbone_generalization}

To test whether the observed gains are specific to TIGER or instead reflect a more general improvement in policy optimization, we repeat the RL-objective comparison on the HSTU backbone under the same reward generation procedure. GBPO maintains competitive recommendation accuracy and consistently stable training across both backbones (Table~\ref{tab:hstu_rl_results}).

Several trends from the TIGER results persist. First, GRPO and GSPO collapse on all three datasets, and DAPO remains unstable on the harder Toys and Sports settings. Second, GPPO becomes substantially more robust after the additional negative-gradient rescaling discussed in Appendix~\ref{app:deriv_gppo}, but still trails GBPO on all three datasets. Third, Dual-Clip PPO improves short-term stability on some settings but does not consistently match GBPO's training stability across datasets. These observations strengthen the claim that the advantage of GBPO comes from the underlying update geometry rather than from a single architecture-specific interaction.

The absolute performance levels of TIGER and HSTU are not identical, which is expected because the two backbones differ in architecture, item representations, and inductive bias. These differences affect the quality of the supervised initialization and the final achievable recommendation ceiling. Despite these differences, GBPO maintains stable training across both backbones. This consistency is important because it suggests that GBPO is improving the optimization geometry of ratio-based RL itself, rather than exploiting an accidental interaction with a specific backbone or tokenizer.
From a practical perspective, this means that practitioners may still need to tune the backbone for the best absolute recommendation accuracy, but the relative robustness of the RL objective appears much less architecture-specific. In that sense, GBPO can be treated as a near drop-in optimization choice for generative recommendation systems, even when the underlying backbone or tokenization scheme changes.

To assess reproducibility, we repeated the Amazon RL comparisons with TIGER and HSTU on a new server using three random seeds (Appendix~\ref{app:amazon_replications}).

\begin{table*}[t]
\centering
\small
\caption{\textbf{RL objective comparison on HSTU backbone.}
All methods share the same HSTU backbone with direct item-ID prediction, candidate generation ($G=32$), frozen DIN reward, and identical optimization settings.
Best results among stable (non-collapsed) methods are in \textbf{bold}.}
\label{tab:hstu_rl_results}
\setlength{\tabcolsep}{0.1pt}
\renewcommand{\arraystretch}{1.4}
\begin{tabular}{l *{12}{>{\centering\arraybackslash}p{30pt}}}
\toprule
& \multicolumn{4}{c}{\textbf{Beauty}}
& \multicolumn{4}{c}{\textbf{Toys}}
& \multicolumn{4}{c}{\textbf{Sports}} \\
\cmidrule(lr){2-5}\cmidrule(lr){6-9}\cmidrule(lr){10-13}
\textbf{RL Method}
& \textbf{H@5} & \textbf{H@10} & \textbf{N@5} & \textbf{N@10}
& \textbf{H@5} & \textbf{H@10} & \textbf{N@5} & \textbf{N@10}
& \textbf{H@5} & \textbf{H@10} & \textbf{N@5} & \textbf{N@10} \\
\midrule
GRPO$^\dagger$~\cite{shao2024deepseekmathpushinglimitsmathematical}
& 0.0405$^\dagger$ & 0.0569$^\dagger$ & 0.0285$^\dagger$ & 0.0338$^\dagger$
& 0.0334$^\dagger$ & 0.0469$^\dagger$ & 0.0234$^\dagger$ & 0.0278$^\dagger$
& 0.0297$^\dagger$ & 0.0455$^\dagger$ & 0.0191$^\dagger$ & 0.0242$^\dagger$ \\
DAPO~\cite{DBLP:conf/nips/YuZZYZYDFLLLLLL25}
& 0.0411 & 0.0593 & 0.0284 & 0.0343
& 0.0286$^\dagger$ & 0.0400$^\dagger$ & 0.0204$^\dagger$ & 0.0240$^\dagger$
& 0.0304$^\dagger$ & 0.0467$^\dagger$ & 0.0196$^\dagger$ & 0.0248$^\dagger$ \\
GSPO$^\dagger$~\cite{zheng2025group}
& 0.0405$^\dagger$ & 0.0569$^\dagger$ & 0.0285$^\dagger$ & 0.0338$^\dagger$
& 0.0334$^\dagger$ & 0.0469$^\dagger$ & 0.0234$^\dagger$ & 0.0278$^\dagger$
& 0.0297$^\dagger$ & 0.0455$^\dagger$ & 0.0191$^\dagger$ & 0.0242$^\dagger$ \\
GPPO~\cite{su2025gppo}
& 0.0464 & 0.0719 & 0.0301 & 0.0384
& 0.0455 & 0.0704 & 0.0296 & 0.0376
& 0.0284 & 0.0436 & 0.0178 & 0.0227 \\
Dual-Clip PPO$^\dagger$~\cite{DBLP:conf/aaai/YeLSSZWYYWGCYZS20}
& 0.0521$^\dagger$ & 0.0771$^\dagger$ & 0.0360$^\dagger$ & 0.0439$^\dagger$
& 0.0524$^\dagger$ & 0.0743$^\dagger$ & 0.0356$^\dagger$ & 0.0426$^\dagger$
& 0.0289$^\dagger$ & 0.0463$^\dagger$ & 0.0188$^\dagger$ & 0.0244$^\dagger$ \\
\midrule
\textbf{GBPO (ours)}
& \textbf{0.0498} & \textbf{0.0741} & \textbf{0.0330} & \textbf{0.0408}
& \textbf{0.0490} & \textbf{0.0717} & \textbf{0.0332} & \textbf{0.0404}
& \textbf{0.0288} & \textbf{0.0450} & \textbf{0.0182} & \textbf{0.0233} \\
\bottomrule
\end{tabular}
\end{table*}

\subsection{Design Choices of GBPO}
\label{subsec:design_choices}

This subsection examines three practical choices that are specific to our training pipeline: whether GBPO should be applied at the token or sequence level, how many positives should be marked in each candidate group, and how large the candidate set should be. These choices directly determine how informative the relative reward signal is and how much variance is injected into the policy update.

\paragraph{Token-level vs.\ sequence-level GBPO}
Section~\ref{subsec:gbpo} presents the token- and sequence-level objectives separately. Our main TIGER experiments use the sequence-level geometric-mean score $P_i$, whose $n_i$th power is the joint SID likelihood. Table~\ref{tab:token_gbpo_results} shows that the two variants are competitive, but the sequence-level version is more consistently robust across datasets, especially on Sports. This supports aligning the normalization granularity with the item-level reward. Both variants admit reference-independent local logit-gradient bounds, with the weaker, length-dependent negative envelope of the sequence form stated explicitly in Eq.~\eqref{eq:gbpo_bound_neg}.

\begin{table}[t]
\centering
\small
\caption{Performance comparison between token-level and sequence-level GBPO.}
\label{tab:token_gbpo_results}
\renewcommand{\arraystretch}{1.4}
\begin{tabular}{ll *{4}{>{\centering\arraybackslash}p{29pt}}}
\toprule
\textbf{Dataset} & \textbf{Variant} & \textbf{H@5} & \textbf{H@10} & \textbf{N@5} & \textbf{N@10} \\
\midrule
\textbf{Beauty} & Sequence & 0.0523 & 0.0753 & 0.0327 & 0.0401 \\
                & Token    & 0.0517 & 0.0773 & 0.0328 & 0.0411 \\
\midrule
\textbf{Toys}   & Sequence & 0.0543 & 0.0792 & 0.0350 & 0.0431 \\
                & Token    & 0.0541 & 0.0800 & 0.0344 & 0.0428 \\
\midrule
\textbf{Sports} & Sequence & 0.0300 & 0.0489 & 0.0187 & 0.0248 \\
                & Token    & 0.0275 & 0.0457 & 0.0170 & 0.0228 \\
\bottomrule
\end{tabular}
\end{table}

\paragraph{Sensitivity to the positive reward threshold $K$.}
Our reward construction marks the top-$K$ DIN-ranked candidates in each group as positives. Table~\ref{tab:candidate_sensitivity} shows that recommendation quality follows a clear bell-shaped trend as $K$ increases. When $K$ is too small, the reward is overly sparse and discards near-miss positives that could provide useful ranking signal; when $K$ is too large, lower-quality candidates are promoted to positives and the contrastive structure becomes diluted. Across all three datasets, $K=5$ offers the best or near-best trade-off, which is why we use it as the default throughout the paper.

\begin{table}[t]
\centering
\small
\caption{Impact of the top-$K$ positive reward threshold on recommendation performance (Top-10).}
\renewcommand{\arraystretch}{1.4}
\setlength{\tabcolsep}{6pt}
\label{tab:candidate_sensitivity}
\begin{tabular}{lcccccc}
\toprule
\textbf{Beauty} & $K=1$ & $K=3$ & $K=5$ & $K=8$ & $K=10$ & $K=16$ \\
\midrule
H@10   & 0.0707 & 0.0753 & 0.0753 & 0.0772 & 0.0764 & 0.0665 \\
N@10 & 0.0400 & 0.0423 & 0.0401 & 0.0396 & 0.0378 & 0.0319 \\
\midrule
\textbf{Toys} & $K=1$ & $K=3$ & $K=5$ & $K=8$ & $K=10$ & $K=16$ \\
\midrule
H@10   & 0.0661 & 0.0755 & 0.0792 & 0.0748 & 0.0751 & 0.0682 \\
N@10 & 0.0384 & 0.0427 & 0.0431 & 0.0393 & 0.0373 & 0.0344 \\
\midrule
\textbf{Sports} & $K=1$ & $K=3$ & $K=5$ & $K=8$ & $K=10$ & $K=16$ \\
\midrule
H@10   & 0.0393 & 0.0455 & 0.0489 & 0.0476 & 0.0474 & 0.0439 \\
N@10 & 0.0223 & 0.0244 & 0.0248 & 0.0237 & 0.0231 & 0.0206 \\
\bottomrule
\end{tabular}
\end{table}

\paragraph{Robustness to reward sparsity.}
To isolate whether the gains from a larger $K$ merely come from denser rewards or also from better use of those rewards, we further compare GBPO with GRPO on Toys under the same varying-$K$ setup. Table~\ref{tab:grpo_vs_gbpo_k} shows that GBPO maintains a substantial lead over GRPO across all settings, with the largest gap appearing near the best-performing region ($K=5$ to $10$). This suggests that GBPO is not only less sensitive to sparse rewards at small $K$, but is also better able to translate increasingly informative positive sets into stable policy improvements as more candidates are marked positive.

\begin{table}[t]
\centering
\small
\renewcommand{\arraystretch}{1.4}
\caption{Comparative performance of GBPO and GRPO under different reward thresholds $K$ (Toys dataset, Top-10).}
\setlength{\tabcolsep}{5pt}
\label{tab:grpo_vs_gbpo_k}
\begin{tabular}{lcccccc}
\toprule
\textbf{Method} & $K=1$ & $K=3$ & $K=5$ & $K=8$ & $K=10$ & $K=16$ \\
\midrule
GBPO H@10   & 0.0661 & 0.0755 & 0.0792 & 0.0748 & 0.0751 & 0.0682 \\
GBPO N@10   & 0.0384 & 0.0427 & 0.0431 & 0.0393 & 0.0373 & 0.0344 \\
\midrule
GRPO H@10   & 0.0599 & 0.0609 & 0.0630 & 0.0637 & 0.0585 & 0.0548 \\
GRPO N@10   & 0.0316 & 0.0322 & 0.0320 & 0.0322 & 0.0300 & 0.0275 \\
\bottomrule
\end{tabular}
\end{table}

\paragraph{Sensitivity to candidate set size $G$.}
The candidate set size controls the granularity of within-group comparisons. Table~\ref{tab:candidate_size_sensitivity} shows that GBPO is relatively robust across a wide range of $G$, but performs best at $G=32$, which is the value used by our training scripts. Smaller groups provide less informative relative ranking structure, whereas excessively large groups increase the number of negatives without proportionally improving the reward signal. Notably, GBPO remains substantially stronger than GRPO even in the smaller-budget regime, which is consistent with the claim that full gradient activity improves sample efficiency.

\begin{table}[t]
\centering
\small
\renewcommand{\arraystretch}{1.4}
\caption{Impact of candidate set size $G$ on recommendation performance (Toys dataset, Top-10).}
\label{tab:candidate_size_sensitivity}
\begin{tabular}{l *{5}{>{\centering\arraybackslash}p{28pt}}}
\toprule
\textbf{Method} & $G=8$ & $G=16$ & $G=32$ & $G=64$ & $G=128$ \\
\midrule
GBPO H@10   & 0.0752 & 0.0776 & {0.0792} & 0.0782 & 0.0767 \\
GBPO N@10 & 0.0409 & 0.0425 & {0.0431} & 0.0425 & 0.0412 \\
\midrule
GRPO H@10            & 0.0523 & 0.0589 & 0.0630 & 0.0630 & 0.0645 \\
GRPO N@10          & 0.0268 & 0.0310 & 0.0320 & 0.0330 & 0.0342 \\
\bottomrule
\end{tabular}
\end{table}

% =========================================================
\subsection{Comparison with Non-RL Post-Training}
\label{subsec:non_rl_post_training}
We compare GBPO, KD, and S-DPO from the same TIGER SFT initialization using seed 42. KD uses a learning rate of $5\times10^{-4}$ and temperature $1$; S-DPO uses a learning rate of $10^{-5}$ and $\beta=1$. Table~\ref{tab:non_rl_post_training} reports their test results under beam size $256$, reusing the corresponding GBPO runs from Appendix~\ref{app:tiger_replication}.

GBPO and KD each lead six of twelve metrics. GBPO leads H@10 on all three datasets and all four metrics on Toys. Both record no collapse/nonfinite-loss event, whereas S-DPO triggers the collapse criterion on all three datasets.

\begin{table}[t]
\centering
\small
\setlength{\tabcolsep}{5pt}
\renewcommand{\arraystretch}{1.15}
\caption{Single-run comparison of GBPO and non-RL post-training methods on TIGER (seed 42). Best values within each dataset are bold.}
\label{tab:non_rl_post_training}
\begin{tabular}{llrrrr}
\toprule
\textbf{Dataset} & \textbf{Method} & \textbf{H@5} & \textbf{H@10} & \textbf{NDCG@5} & \textbf{NDCG@10} \\
\midrule
\textbf{Beauty} & \textbf{GBPO} & 0.0511 & \textbf{0.0774} & 0.0325 & 0.0410\\
 & KD & \textbf{0.0529} & 0.0766 & \textbf{0.0368} & \textbf{0.0444}\\
 & S-DPO$^{\dagger}$ & 0.0440 & 0.0694 & 0.0296 & 0.0377\\
\midrule
\textbf{Toys} & \textbf{GBPO} & \textbf{0.0541} & \textbf{0.0782} & \textbf{0.0350} & \textbf{0.0428}\\
 & KD & 0.0506 & 0.0745 & 0.0349 & 0.0426\\
 & S-DPO$^{\dagger}$ & 0.0347 & 0.0559 & 0.0222 & 0.0289\\
\midrule
\textbf{Sports} & \textbf{GBPO} & 0.0301 & \textbf{0.0475} & 0.0188 & 0.0244\\
 & KD & \textbf{0.0315} & 0.0470 & \textbf{0.0209} & \textbf{0.0259}\\
 & S-DPO$^{\dagger}$ & 0.0221 & 0.0367 & 0.0141 & 0.0188\\
\bottomrule
\end{tabular}
\end{table}

\subsection{Reward Mapping and Reward-Model Sensitivity}
\label{subsec:reward_robustness}
We assess reward mapping and teacher architecture on TIGER, retaining the student and candidate interface. Appendix~\ref{app:extended_reward_mapping} and Appendix~\ref{app:extended_teacher} report all six methods on three datasets.

\paragraph{Continuous DIN scores.}
With continuous DIN sigmoid scores instead of binary rewards, GBPO leads nine of twelve metrics and records no collapse/nonfinite-loss event. All fifteen competing runs trigger the collapse criterion (D). This tests a qualitatively different reward mapping inspired by score-based recommendation rewards~\cite{DBLP:journals/corr/abs-2510-24431}.

\paragraph{Alternative ranking teacher.}
With a separately trained SASRec-style teacher and binary rewards, GBPO leads nine of twelve metrics, including all four metrics on Toys and Sports and H@10 on Beauty, and records no collapse/nonfinite-loss event on any dataset. On Beauty, GSPO leads H@5 and NDCG@5, whereas Dual-Clip leads NDCG@10. All fifteen competing runs trigger the collapse criterion (D). Accuracy rankings vary with the teacher, while GBPO retains its stopping-stability advantage.

\subsection{A Non-Amazon Domain: Steam}
\label{subsec:steam}
We add a benchmark from public Steam reviews\footnote{\url{https://mcauleylab.ucsd.edu/public_datasets/data/steam/}}. Metadata filtering, user sampling, and iterative 5-core filtering yield 29,357 user identifiers, 5,734 games, and 354,088 interactions. We use chronological leave-one-out splits with deterministic handling of date ties, and construct SIDs from game titles and genres. All six RL objectives achieve similar gains over SFT across three seeds (Table~\ref{tab:steam_results}). Only GBPO avoids performance-collapse and nonfinite-loss events in every Steam run.

\begin{table}[t]
\centering
\small
\setlength{\tabcolsep}{3pt}
\renewcommand{\arraystretch}{1.2}
\caption{Steam results with the TIGER backbone and DIN rewards. SFT is the fixed initialization; RL results are mean $\pm$ sample standard deviation over seeds 42--44. The highest RL means are bold; the second-highest are underlined.}
\label{tab:steam_results}
\begin{tabular}{lccccr}
\toprule
\textbf{Method} & \textbf{H@5} & \textbf{H@10} & \textbf{NDCG@5} & \textbf{NDCG@10} & \textbf{D/N} \\
\midrule
SFT & 0.04561 & 0.07589 & 0.02901 & 0.03874 &  \\
GRPO & $0.05281\pm0.00037$ & $0.08825\pm0.00109$ & $0.03326\pm0.00040$ & $0.04466\pm0.00060$ & 3/0 \\
DAPO & $0.05264\pm0.00073$ & $0.08716\pm0.00125$ & $0.03306\pm0.00056$ & $0.04415\pm0.00074$ & 3/0 \\
GSPO & $\mathbf{0.05577}\pm\mathbf{0.00036}$ & $\mathbf{0.09080}\pm\mathbf{0.00093}$ & $\mathbf{0.03475}\pm\mathbf{0.00017}$ & $\mathbf{0.04598}\pm\mathbf{0.00026}$ & 3/0 \\
GPPO & $0.05382\pm0.00152$ & $0.08920\pm0.00061$ & $0.03348\pm0.00125$ & $0.04485\pm0.00079$ & 3/0 \\
Dual-Clip & $0.05419\pm0.00079$ & $0.08779\pm0.00101$ & $0.03392\pm0.00057$ & $0.04475\pm0.00052$ & 3/2 \\
\textbf{GBPO} & $\underline{0.05509\pm0.00084}$ & $\underline{0.09011\pm0.00057}$ & $\underline{0.03395\pm0.00068}$ & $\underline{0.04519\pm0.00052}$ & 0/0 \\
\bottomrule
\end{tabular}
\end{table}

\subsection{Optimization Stability and Training Dynamics}
\label{subsec:stability_dynamics}

\begin{figure}[t]
\centering
\includegraphics[width=0.8\linewidth]{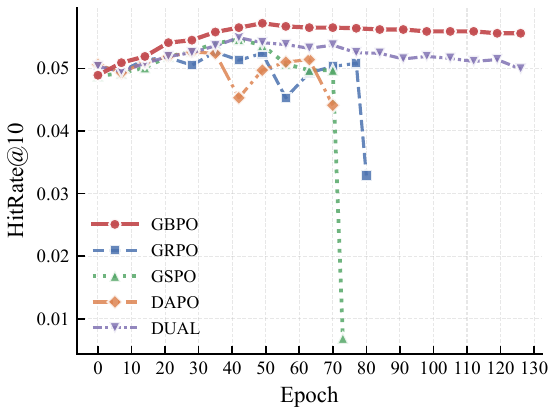}
%\vspace{-4pt}
\caption{Comparison of training dynamics across RL objectives in Amazon Sports.}
\label{fig:dynamics_lr5e-5}
\Description{Validation HitRate@10 over training steps comparing various RL objectives.}
%\vspace{-6pt}
\end{figure}

\begin{figure}[t]
\centering
\includegraphics[width=0.8\linewidth]{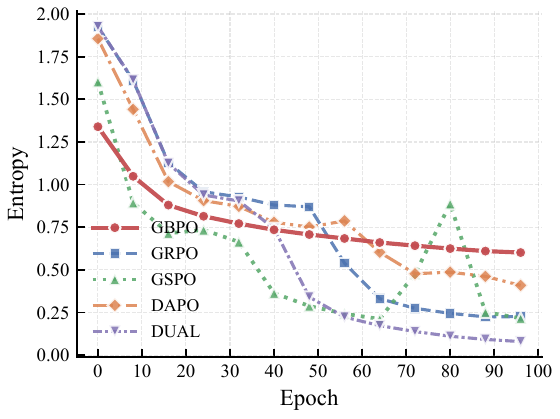}
%\vspace{-4pt}
\caption{Comparison of entropy evolution across RL methods in Amazon Sports.}
\Description{Entropy evolution over training epochs for different reinforcement learning methods.}
\label{fig:grad_matched}
%\vspace{-6pt}
\end{figure}

\begin{figure}[htbp]
    \centering
    \includegraphics[width=0.8\linewidth]{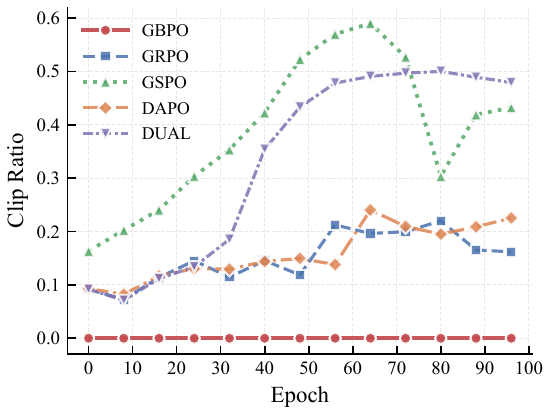}
    \Description{Plot showing the proportion of samples with discarded gradients across training for various RL methods, illustrating GBPO's full sample utilization.}
    \caption[]{\textbf{Proportion of Samples with Discarded Gradients across RL Methods in Amazon Sports.} Clipping-based baselines exhibit a consistently non-zero discarded ratio, indicating under-utilization of samples,
    whereas GBPO stays near zero by design (no clip), enabling all sampled tokens to contribute learning signal.}
    \label{fig:gradient_region_clip}
\end{figure}

\begin{figure}[htbp]
    \centering
    \includegraphics[width=0.8\linewidth]{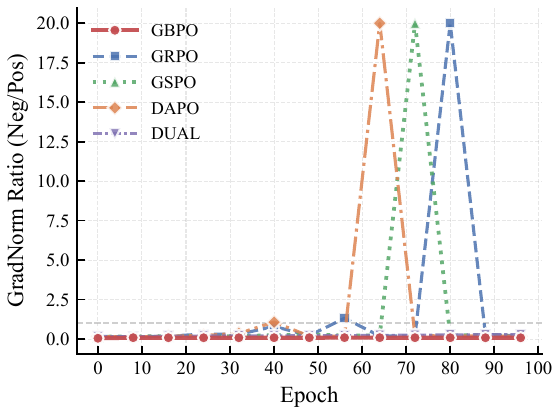}
    \Description{Plot of the gradient-norm ratio between negative and positive advantage updates, showing stability differences among RL methods.}
    \caption[]{\textbf{Gradient-Norm Ratio (Negative vs.\ Positive Advantage) across RL Methods in Amazon Sports.} The ratio $g_{\mathrm{neg}}/g_{\mathrm{pos}}$ quantifies the dominance of negative updates (clipped to 20 for visualization, which is sufficient to show the dominance of negative gradients). Spikes in this ratio often precede training collapse.}
    \label{fig:g_rl_neg_epoch_avg}
\end{figure}

\paragraph{Design-to-evidence mapping.}
We analyze optimization behaviors using diagnostics that directly reflect the three issues identified in Section~\ref{subsec:diagnosis}:
(1) \emph{sample utilization} via discarded-gradient ratios (Fig.~\ref{fig:gradient_region_clip}),
(2) \emph{negative-side stability} via the negative-to-positive gradient ratio and its alignment with collapse behavior (Fig.~\ref{fig:g_rl_neg_epoch_avg} and Fig.~\ref{fig:dynamics_lr5e-5}),
and (3) \emph{confidence-aware hard-example emphasis} via long-horizon validation retention and entropy evolution (Fig.~\ref{fig:dynamics_lr5e-5} and Fig.~\ref{fig:grad_matched}).

\paragraph{Why these diagnostics are sufficient.}
These diagnostics are chosen to match the three structural properties in our analysis rather than to provide an arbitrary collection of training curves.
The discarded-gradient ratio is a direct empirical proxy for sample utilization: when the surrogate enters a flat clipped region, the sampled token contributes zero gradient, so the corresponding update is counted as discarded. A persistently non-zero discarded ratio therefore indicates that a portion of sampled trajectories is systematically ignored, exactly matching the dead-gradient pathology analyzed in Section~\ref{subsec:diagnosis}.
The ratio $g_{\mathrm{neg}}/g_{\mathrm{pos}}$ is informative because it reveals when training becomes increasingly dominated by negative-advantage updates. In our experiments, spikes in this ratio closely coincide with the onset of sharp validation deterioration, suggesting that negative-side updates are a primary driver of instability. This makes $g_{\mathrm{neg}}/g_{\mathrm{pos}}$ a useful diagnostic for identifying negative-side collapse behavior during optimization.
Finally, the entropy trajectory helps interpret confidence-aware weighting. If an objective keeps allocating substantial update mass to already confident cases, policy entropy tends to decay too quickly and training enters an over-confident regime early. In contrast, a method that increasingly suppresses updates on high-confidence positives and already-suppressed negatives should exhibit fast early structure formation together with slower long-horizon entropy decay. This is precisely the behavior shown by GBPO in Fig.~\ref{fig:grad_matched}, which is why the entropy curve is a useful diagnostic for the confidence-weighting property rather than just a generic optimization statistic.

\paragraph{Stability protocol.}
Note that GPPO is excluded from the training dynamics and gradient diagnostic plots (Figs.~\ref{fig:dynamics_lr5e-5}--\ref{fig:g_rl_neg_epoch_avg}) because it collapses within the first epoch without the negative-gradient scaling described in Appendix~\ref{app:deriv_gppo}. This rapid collapse further motivates the need for sign-symmetric gradient boundedness, which GBPO achieves by design. The GPPO results reported in Table~\ref{tab:rl_results} use the rescaled variant (negative-advantage gradients scaled by 0.01) to provide a stronger baseline comparison.

\paragraph{(1) Empirical validation of sample utilization.}
Clipping-based surrogates can enter saturation regions where sampled tokens contribute no gradient.
Figure~\ref{fig:gradient_region_clip} quantifies this effect: clipping/truncation-based baselines exhibit a consistently non-zero discarded-gradient ratio, indicating under-utilization of samples.
By construction, GBPO remains fully gradient-active and keeps the discarded ratio zero, ensuring that all sampled tokens can contribute learning signal.

\paragraph{(2) Bounded negative updates and stability.}
Figure~\ref{fig:dynamics_lr5e-5} plots validation HitRate@10 throughout training; collapsing baselines (e.g., GRPO/GSPO/DAPO) suffer from sudden performance degradation that does not recover.
To directly connect the collapse to optimization dynamics, Figure~\ref{fig:g_rl_neg_epoch_avg} reports the ratio between negative-advantage and positive-advantage gradient norms, i.e., $g_{\mathrm{neg}}/g_{\mathrm{pos}}$. Strikingly, the spike location in $g_{\mathrm{neg}}/g_{\mathrm{pos}}$ aligns with the exact stage where validation HitRate@10 sharply deteriorates, indicating that optimization becomes dominated by negative-side updates.
This observation is consistent with our analysis that ratio-based objectives can amplify negative updates and destabilize training.
In contrast, GBPO enforces sign-symmetric boundedness and suppresses already-suppressed negatives, preventing sudden dominance of negative updates and yielding substantially more stable trajectories.

\paragraph{(3) Confidence-aware weighting and long-horizon retention.}
Beyond bounding updates, stability also depends on \emph{where} the learning signal is allocated.
Figure~\ref{fig:dynamics_lr5e-5} shows that GBPO maintains a high validation plateau for longer, whereas Dual-Clip PPO---although it stabilizes training by constraining negative-side updates---still exhibits a gradual downward drift.
A plausible explanation is that GBPO further introduces confidence-aware weighting under the \emph{current} policy: as training progresses, it increasingly attenuates updates on already high-confidence cases (high-confidence positives / already-suppressed negatives), preventing over-penalization and distribution collapse.
Figure~\ref{fig:grad_matched} shows a rapid initial entropy decrease followed by slower decay for GBPO. These entropy dynamics are consistent with confidence-aware weighting.

\paragraph{Other datasets.}
We provide the same training-dynamics and gradient diagnostics plots on Beauty and Toys in Appendix~\ref{app:other_dataset_metrics}.

\paragraph{Summary of experimental findings.}
Across the evaluated datasets, GBPO achieves competitive recommendation accuracy and demonstrates more stable training than the compared RL objectives.
These gains are consistent with our analysis: GBPO avoids dead-gradient truncation, bounds both update signs, and applies confidence-aware weighting to emphasize hard cases.

\begin{table*}[t]
\centering
\small
\caption{\textbf{RL objective comparison on TIGER backbone.}
All methods report HitRate (H) and NDCG (N). They use constrained SID decoding, beam-search candidate generation ($G=32$ beams), frozen DIN reward, and identical optimization settings unless stated otherwise.}
\label{tab:rl_results}
\setlength{\tabcolsep}{0.1pt}
\renewcommand{\arraystretch}{1.4}
\begin{tabular}{l *{12}{>{\centering\arraybackslash}p{30pt}}}
\toprule
& \multicolumn{4}{c}{\textbf{Beauty}} 
& \multicolumn{4}{c}{\textbf{Toys}} 
& \multicolumn{4}{c}{\textbf{Sports}} \\
\cmidrule(lr){2-5}\cmidrule(lr){6-9}\cmidrule(lr){10-13}
\textbf{RL Method}
& \textbf{H@5} & \textbf{H@10} & \textbf{N@5} & \textbf{N@10}
& \textbf{H@5} & \textbf{H@10} & \textbf{N@5} & \textbf{N@10}
& \textbf{H@5} & \textbf{H@10} & \textbf{N@5} & \textbf{N@10} \\
\midrule
GRPO~\cite{shao2024deepseekmathpushinglimitsmathematical} 
& 0.0469 & 0.0719 & 0.0298 & 0.0379  & 0.0385 & 0.0630 & 0.0242 & 0.0320  & 0.0258 & 0.0402 & 0.0160 & 0.0206 \\
DAPO~\cite{DBLP:conf/nips/YuZZYZYDFLLLLLL25}
& 0.0463 & 0.0723 & 0.0300 & 0.0384  & 0.0390 & 0.0636 & 0.0250 & 0.0329  & 0.0255 & 0.0394 & 0.0158 & 0.0203 \\
GSPO~\cite{zheng2025group} 
& \underline{0.0491} & \underline{0.0747} & 0.0315 & \underline{0.0397}  
& \underline{0.0479} & \underline{0.0715} & \underline{0.0308} & \underline{0.0384}  
& \underline{0.0285} & 0.0444 & \underline{0.0180} & \underline{0.0232} \\
GPPO~\cite{su2025gppo} 
& 0.0454 & 0.0729 & 0.0287 & 0.0375  & 0.0429 & 0.0667 & 0.0269 & 0.0346  & 0.0275 & \underline{0.0447} & 0.0172 & 0.0228 \\
Dual-Clip PPO~\cite{DBLP:conf/aaai/YeLSSZWYYWGCYZS20}
& 0.0486 & 0.0732 & \underline{0.0317} & 0.0396  
& 0.0439 & 0.0668 & 0.0280 & 0.0354  
& 0.0265 & 0.0432 & 0.0169 & 0.0223 \\
\midrule
\textbf{GBPO (ours)} 
& \textbf{0.0523} & \textbf{0.0753} & \textbf{0.0327} & \textbf{0.0401}
& \textbf{0.0543} & \textbf{0.0792} & \textbf{0.0350} & \textbf{0.0431}
& \textbf{0.0300} & \textbf{0.0489} & \textbf{0.0187} & \textbf{0.0248} \\
\bottomrule
\end{tabular}
\end{table*}

\section{Ethics Statement}
In this work, we conduct experiments on open-source benchmark datasets. All datasets are publicly available from previous works while maintaining anonymity. Our training process relies exclusively on aggregated user behavioral sequences, item-level textual content, and non-sensitive metadata, without accessing or processing any personally identifiable information. 

\section{Conclusion}
In this work, we studied reinforcement learning for generative recommendation under semantic-ID (SID) sequence generation and addressed the stability challenges inherent in ratio-based policy optimization. Through a unified logit-gradient view, we identified three structural issues shared by widely used ratio-based objectives in this setting: (i) clipped surrogates can create dead-gradient regions, resulting in under-utilization of samples; (ii) negative-advantage updates can be effectively unbounded as importance ratios grow, allowing a small subset of negatives to dominate optimization and destabilize training; and (iii) ratio-based control lacks sensitivity to absolute policy confidence, meaning updates on high-confidence cases may fail to vanish when they are driven by ratios to the behavior policy rather than absolute confidence.

Motivated by this diagnosis, we proposed Gradient-Bounded Policy Optimization (GBPO). GBPO introduces a stop-gradient, confidence-aware normalization that is fully gradient-active, explicitly bounds both positive and negative updates in a sign-symmetric manner, and ties update magnitude to the current policy confidence. This design naturally emphasizes hard and informative examples while suppressing overly confident ones, improving optimization stability and generation quality. These results highlight GBPO's training-stability advantage among the evaluated RL objectives, alongside competitive recommendation accuracy.

More broadly, our results suggest that stable RL for generative recommendation requires more than simply clipping policy ratios more aggressively. What matters is the joint geometry of the update: all sampled cases should remain trainable, negative-side corrections should be explicitly controlled, and optimization effort should decay as the policy becomes confident. In this sense, GBPO should be viewed not only as a new objective, but also as a concrete design principle for adapting post-training RL methods from open-ended language generation to structured recommendation settings with constrained action spaces.

Future research may explore promising directions, including extending GBPO to richer feedback signals (e.g., long-horizon user satisfaction, diversity, and fairness), as well as adapting the framework to more complex constrained-generation settings while preserving stability and efficiency.

\section{Limitations}
While GBPO improves the optimization behavior of ratio-based policy learning for SID-based generative recommendation, several limitations remain.

\paragraph{RL cost and engineering overhead.}
GBPO is a reinforcement learning method that repeatedly performs a generate--evaluate--update loop: generating candidate items with the reference policy, scoring them with an external evaluator, and updating model parameters.
This can be computationally expensive for large models, especially when constrained decoding is used and multiple candidates per context are required for reliable relative advantages.
As a result, GBPO may have lower training throughput than supervised learning and can be sensitive to practical choices such as sampling temperature, candidate set size, and decoding constraints, which affect both the learning signal and the exploration--exploitation trade-off.
In addition, the engineering stack is more involved than standard supervised recommendation: one must maintain the generative backbone, the constrained decoder, the candidate generator, the frozen reward model, and logging for RL-specific diagnostics. This complexity raises the barrier to reproduction and makes deployment-time debugging more difficult than in conventional next-item training pipelines.

\paragraph{Dependence on reward quality.}
GBPO relies on an external reward signal (e.g., a ranking/scoring model) that is treated as a black box.
Any bias, misspecification, or distribution shift in this reward component can cap the achievable gains and may induce reward hacking behaviors.
Although GBPO stabilizes policy-gradient updates, it does not by itself guarantee that the learned policy aligns with the true user utility beyond what the reward captures.
Improving reward reliability (e.g., calibration, robustness, and out-of-distribution generalization) remains crucial.
This limitation is particularly important in recommendation because reward models are often trained on historical feedback that is itself affected by exposure bias, popularity effects, and previous serving policies. A method may therefore become better at optimizing the learned evaluator while still inheriting blind spots in the evaluator's notion of utility. In that sense, GBPO addresses optimization pathologies downstream of the reward, but does not solve the upstream problem of reward specification.

\paragraph{Deployment with online feedback.}
Extending GBPO to online learning introduces additional challenges.
Real user feedback is often delayed, noisy, and non-stationary, and the action space under constrained SID generation can change with catalog dynamics.
These factors complicate stable policy optimization and typically require counterfactual correction, careful traffic allocation, and continuous monitoring to avoid regressions.
Moreover, our experiments are entirely offline and focus on next-item recommendation accuracy. They do not evaluate broader deployment objectives such as diversity, novelty, fairness, user fatigue, or long-term ecosystem effects. In an online system, these objectives are often coupled, so a practically useful RL method may need to optimize a multi-objective reward rather than a single ranking-oriented signal.

\paragraph{Generality beyond the current setting.}
Our study centers on SID-based autoregressive recommendation with constrained generation and groupwise relative rewards. Although the empirical evidence spans two strong generative backbones, the conclusions may not transfer unchanged to settings with different action parameterizations, denser rewards, or substantially longer generation horizons. In particular, tasks with richer textual actions, mixed discrete-continuous action spaces, or delayed sequence-level credit assignment may require different normalization or exploration mechanisms.

Overall, GBPO addresses key \emph{optimization} pathologies of ratio-based objectives (e.g., dead-gradient regions, unbounded negative updates, and lack of sensitivity to absolute policy confidence), but it does not remove fundamental bottlenecks related to sampling cost, reward quality, deployment realism, and task generality.
Future work includes more data-efficient training, principled exploration under constrained decoding, stronger reward modeling, and broader evaluation protocols that connect offline optimization stability to online recommendation quality. More specifically, an important next step is to establish a clearer link between offline stability diagnostics and online business or user-centric metrics, so that optimization improvements can be evaluated not only in terms of training behavior, but also in terms of practical impact after deployment.

%\begin{acks}
%To Robert, for the bagels and explaining CMYK and color spaces.
%\end{acks}

\bibliographystyle{ACM-Reference-Format}
\bibliography{sample-base}

\appendix
\captionsetup{hypcap=false}

\section{Logit-Gradient Derivations}
\label{app:grad_derivations_full}

\begingroup\small

\subsection{Softmax Derivatives}
\label{app:softmax_identities}
Let $z\in\mathbb{R}^{|V|}$ be logits and $\pi=\mathrm{softmax}(z)$ with
$\pi_k=\exp(z_k)/\sum_m \exp(z_m)$.
Then
\begin{equation}
\label{eq:softmax_jacobian_app}
\frac{\partial \pi_k}{\partial z_j}
=
\pi_k\bigl(\mathbb{I}[j=k]-\pi_j\bigr),
\end{equation}
and
\begin{equation}
\label{eq:logsoftmax_grad_app}
\frac{\partial \log \pi_k}{\partial z_j}
=
\mathbb{I}[j=k]-\pi_j.
\end{equation}

\paragraph{Sufficiency of the selected-token gradient.}
For a loss term $\ell = -g(\pi_k)$ depending on a specific selected token $k$, the logit gradient w.r.t.\ any $z_j$ is $\partial \ell / \partial z_j = -g'(\pi_k) \pi_k (\mathbb{I}[j=k] - \pi_j)$. Summing the gradients of all non-selected tokens ($j \neq k$) yields:
\begin{equation}
\label{eq:grad_sufficiency_proof}
\sum_{j \neq k} \frac{\partial \ell}{\partial z_j} = g'(\pi_k) \pi_k \sum_{j \neq k} \pi_j = g'(\pi_k) \pi_k (1 - \pi_k) = - \frac{\partial \ell}{\partial z_k}.
\end{equation}
This identity shows that the gradient of the selected logit $z_k$ is exactly balanced by the sum of all other gradients, satisfying $\sum_{j} \partial \ell / \partial z_j = 0$. Consequently, the magnitude $|\partial \ell / \partial z_k|$ fully characterizes the update intensity for the entire logit vector, justifying our focus on the selected-token logit gradient throughout Section~\ref{subsec:diagnosis} and Section~\ref{subsec:gbpo}.

\subsection{Template}
\label{app:template_loss_to_grad}
Fix a trajectory $o_i$ and time step $t$ with selected token $o_{i,t}$ and context $\mathrm{ctx}_{i,t}$.
Denote $\pi_{i,t,j}\triangleq \pi_\theta(j\mid\mathrm{ctx}_{i,t})$ and logit $z_{i,t,j}$.
Define the token-level ratio
\begin{equation}
\label{eq:ratio_def_app}
r_{i,t}(\theta)\triangleq
\frac{\pi_\theta(o_{i,t}\mid \mathrm{ctx}_{i,t})}{\pi_{\mathrm{old}}(o_{i,t}\mid \mathrm{ctx}_{i,t})},
\end{equation}
where $\pi_{\mathrm{old}}$ is treated as constant w.r.t.\ $\theta$.

\paragraph{Step T1: derivative of the ratio.}
Using \eqref{eq:softmax_jacobian_app} with $k=o_{i,t}$,
\begin{equation}
\label{eq:dr_dz_app_compact}
\frac{\partial r_{i,t}}{\partial z_{i,t,j}}
=
r_{i,t}\bigl(\mathbb{I}[j=o_{i,t}]-\pi_{i,t,j}\bigr).
\end{equation}

\paragraph{Step T2: generic ratio-based surrogate.}
Consider a per-token loss term of the form
\begin{equation}
\label{eq:generic_loss_term_app}
\ell_{i,t}(\theta)= -\, g_{i,t}\!\big(r_{i,t}(\theta)\big),
\end{equation}
where $g_{i,t}(\cdot)$ is a (possibly clipped/piecewise) scalar function.
By chain rule and \eqref{eq:dr_dz_app_compact},
\begin{equation}
\label{eq:dl_dz_template_app}
\frac{\partial \ell_{i,t}}{\partial z_{i,t,j}}
=
-\frac{\partial g_{i,t}}{\partial r_{i,t}}\,
r_{i,t}\bigl(\mathbb{I}[j=o_{i,t}]-\pi_{i,t,j}\bigr).
\end{equation}
Thus, deriving logit gradients reduces to computing $\partial g_{i,t}/\partial r_{i,t}$.

% ============================================================
\subsection{GRPO}
\label{app:deriv_grpo}
Let $\hat A_i$ be the (group-normalized) advantage shared across tokens in $o_i$.

\paragraph{Step G1: per-token loss.}
\begin{equation}
\label{eq:grpo_loss_token_app}
\ell^{\mathrm{GRPO}}_{i,t}(\theta)
\triangleq
-\min\!\Bigl(
r_{i,t}\hat A_i,\;
\operatorname{clip}(r_{i,t},1-\epsilon,1+\epsilon)\hat A_i
\Bigr).
\end{equation}

\paragraph{Step G2: $\partial g/\partial r$ (indicator).}
Let $u(r)=r\hat A_i$ and $v(r)=\operatorname{clip}(r,1-\epsilon,1+\epsilon)\hat A_i$.
 Define the unclipped-branch indicator
\begin{equation}
\label{eq:grpo_indicator_char_app}
\mathbb{I}^{\mathrm{GRPO}}_{i,t}=
\begin{cases}
\mathbb{I}\!\big[r_{i,t}\le 1+\epsilon\big], & \hat A_i>0,\\
\mathbb{I}\!\big[r_{i,t}\ge 1-\epsilon\big], & \hat A_i<0.
\end{cases}
\end{equation}
Then $\frac{\partial}{\partial r}\min(u,v)=\mathbb{I}^{\mathrm{GRPO}}_{i,t}\hat A_i$ and
\begin{equation}
\label{eq:grpo_dgdr_app}
\frac{\partial g^{\mathrm{GRPO}}_{i,t}}{\partial r_{i,t}}
=
\mathbb{I}^{\mathrm{GRPO}}_{i,t}\hat A_i.
\end{equation}

\paragraph{Step G3: logit gradient.}
Plugging \eqref{eq:grpo_dgdr_app} into \eqref{eq:dl_dz_template_app} yields
\begin{equation}
\label{eq:grpo_logit_grad_app}
\frac{\partial \ell^{\mathrm{GRPO}}_{i,t}}{\partial z_{i,t,j}}
=
\mathbb{I}^{\mathrm{GRPO}}_{i,t}\,\hat A_i\,r_{i,t}\,
\bigl(\pi_{i,t,j}-\mathbb{I}[j=o_{i,t}]\bigr).
\end{equation}

% ============================================================
\subsection{DAPO}
\label{app:deriv_dapo}

\paragraph{Step D1: per-token loss.}
\begin{equation}
\label{eq:dapo_loss_token_app}
\ell^{\mathrm{DAPO}}_{i,t}(\theta)
\triangleq
-\min\!\Bigl(
r_{i,t}\hat A_{i,t},\;
\operatorname{clip}(r_{i,t},1-\epsilon_{\mathrm{low}},1+\epsilon_{\mathrm{high}})\hat A_{i,t}
\Bigr).
\end{equation}

\paragraph{Step D2: $\partial g/\partial r$ (indicator).}
Define
\begin{equation}
\label{eq:dapo_indicator_char_app}
\mathbb{I}^{\mathrm{DAPO}}_{i,t}=
\begin{cases}
\mathbb{I}\!\big[r_{i,t}\le 1+\epsilon_{\mathrm{high}}\big], & \hat A_{i,t}>0,\\
\mathbb{I}\!\big[r_{i,t}\ge 1-\epsilon_{\mathrm{low}}\big], & \hat A_{i,t}<0.
\end{cases}
\end{equation}
Then
\begin{equation}
\label{eq:dapo_dgdr_app}
\frac{\partial g^{\mathrm{DAPO}}_{i,t}}{\partial r_{i,t}}
=
\mathbb{I}^{\mathrm{DAPO}}_{i,t}\hat A_{i,t}.
\end{equation}

\paragraph{Step D3: logit gradient.}
\begin{equation}
\label{eq:dapo_logit_grad_app}
\frac{\partial \ell^{\mathrm{DAPO}}_{i,t}}{\partial z_{i,t,j}}
=
\mathbb{I}^{\mathrm{DAPO}}_{i,t}\,\hat A_{i,t}\,r_{i,t}\,
\bigl(\pi_{i,t,j}-\mathbb{I}[j=o_{i,t}]\bigr).
\end{equation}

% ============================================================
\subsection{GSPO}
\label{app:deriv_gspo}

\paragraph{Step S1: define $\tilde r_i$ and loss.}
\begin{equation}
\label{eq:gspo_si_app}
\tilde r_i(\theta)
\triangleq
\exp\!\left(\frac{1}{|o_i|}\sum_{\tau=1}^{|o_i|}\log r_{i,\tau}\right),
\end{equation}
and
\begin{equation}
\label{eq:gspo_loss_app}
\ell^{\mathrm{GSPO}}_{i}(\theta)
\triangleq
-\min\!\Bigl(
\tilde r_i A_i^{b},\;
\operatorname{clip}(\tilde r_i,1-\epsilon,1+\epsilon) A_i^{b}
\Bigr).
\end{equation}

\paragraph{Step S2: $\partial g/\partial \tilde r$ (sequence indicator).}
Define
\begin{equation}
\label{eq:gspo_indicator_char_app}
\mathbb{I}^{\mathrm{GSPO}}_{i}=
\begin{cases}
\mathbb{I}\!\big[\tilde r_i\le 1+\epsilon\big], & A_i^{b}>0,\\
\mathbb{I}\!\big[\tilde r_i\ge 1-\epsilon\big], & A_i^{b}<0.
\end{cases}
\end{equation}
Then $\frac{\partial}{\partial \tilde r}\min(\tilde r A,\mathrm{clip}(\tilde r)A)=\mathbb{I}^{\mathrm{GSPO}}_i A_i^b$, so
\begin{equation}
\label{eq:gspo_dgds_app}
\frac{\partial g^{\mathrm{GSPO}}_{i}}{\partial \tilde r_i}
=
\mathbb{I}^{\mathrm{GSPO}}_{i}A_i^b.
\end{equation}

\paragraph{Step S3: $\partial \tilde r_i/\partial z_{i,t,j}$.}
From \eqref{eq:gspo_si_app} and \eqref{eq:logsoftmax_grad_app},
\begin{equation}
\label{eq:ds_dz_app_compact}
\frac{\partial \tilde r_i}{\partial z_{i,t,j}}
=
\frac{\tilde r_i}{|o_i|}\bigl(\mathbb{I}[j=o_{i,t}]-\pi_{i,t,j}\bigr).
\end{equation}

\paragraph{Step S4: logit gradient.}
Chain rule gives
\begin{equation}
\label{eq:gspo_logit_grad_app}
\frac{\partial \ell^{\mathrm{GSPO}}_{i}}{\partial z_{i,t,j}}
=
\mathbb{I}^{\mathrm{GSPO}}_{i}\,A_i^{b}\,
\frac{\tilde r_i}{|o_i|}\,
\bigl(\pi_{i,t,j}-\mathbb{I}[j=o_{i,t}]\bigr).
\end{equation}

% ============================================================
\subsection{GPPO}
\label{app:deriv_gppo}

\paragraph{Step P1: per-token loss and stop-grad clip.}
\begin{equation}
\label{eq:gppo_loss_token_app}
\ell^{\mathrm{GPPO}}_{i,t}(\theta)
\triangleq
-\min\!\Bigl(r_{i,t}\hat A_{i,t},\;\bar r_{i,t}\hat A_{i,t}\Bigr),
\end{equation}
with
\begin{equation}
\label{eq:gppo_rbar_app}
\bar r_{i,t}(\theta)
=
\mathrm{clip}\!\left(
r_{i,t},\,
(1-\epsilon_{\mathrm{low}})\frac{r_{i,t}}{\mathrm{sg}(r_{i,t})},\,
(1+\epsilon_{\mathrm{high}})\frac{r_{i,t}}{\mathrm{sg}(r_{i,t})}
\right).
\end{equation}

\paragraph{Step P2: effective $\partial g/\partial r$ (piecewise constant in saturation).}
Because $\mathrm{sg}(\cdot)$ treats $r_{i,t}$ as a constant in the denominators,
$\bar r_{i,t}$ is piecewise linear in $r_{i,t}$ and yields a piecewise-constant slope.
The resulting logit-gradient can be written compactly as
\begin{equation}
\label{eq:gppo_logit_grad_app}
\frac{\partial \ell^{\mathrm{GPPO}}_{i,t}}{\partial z_{i,t,j}}
=\kappa^{\mathrm{GPPO}}_{i,t}\,
\bigl(\pi_{i,t,j}-\mathbb{I}[j=o_{i,t}]\bigr),
\end{equation}
where $\kappa^{\mathrm{GPPO}}_{i,t}$ is the \emph{piecewise} coefficient given in Sec.~\ref{app:piecewise_coeffs}. In practice, we additionally scale the negative-advantage gradients of GPPO by 0.01 to improve its optimization stability; without this scaling, GPPO collapses significantly faster. This extra tuning is not part of the original GPPO formulation but is applied to strengthen the baseline comparison.

% ============================================================
\subsection{Dual-Clip PPO}
\label{app:deriv_dualppo}

\paragraph{Step U1: per-token loss.}
Let $L^{\mathrm{clip}}_{i,t}(\theta)$ denote the standard PPO clipped surrogate
(as in \eqref{eq:grpo_loss_token_app} but without the leading minus sign).
Define
\begin{equation}
\label{eq:dual_loss_token_app}
\ell^{\mathrm{DUAL}}_{i,t}(\theta)
\triangleq
-\,
\begin{cases}
L^{\mathrm{clip}}_{i,t}(\theta), & \hat A_{i,t}\ge 0,\\[2pt]
\max\!\big(L^{\mathrm{clip}}_{i,t}(\theta),\,c\,\hat A_{i,t}\big), & \hat A_{i,t}<0,
\end{cases}
\end{equation}
where the supplied experiments use $c=3$.

\paragraph{Step U2: floor indicator and logit gradient.}
When $\hat A_{i,t}<0$, the constant floor $c\hat A_{i,t}$ is independent of $\theta$.
Thus the gradient is zero whenever the floor is active.
The resulting logit-gradient can be written as
\begin{equation}
\label{eq:dual_logit_grad_app}
\frac{\partial \ell^{\mathrm{DUAL}}_{i,t}}{\partial z_{i,t,j}}
=
\kappa^{\mathrm{DUAL}}_{i,t}\,
\bigl(\pi_{i,t,j}-\mathbb{I}[j=o_{i,t}]\bigr),
\end{equation}
where $\kappa^{\mathrm{DUAL}}_{i,t}$ is the piecewise coefficient given in
Sec.~\ref{app:piecewise_coeffs}.

% ============================================================
\subsection{Token-Level GBPO}
\label{app:deriv_gbpo}

\paragraph{Step B1: objective and normalization.}
Recall the token-level objective in Eq.~\eqref{eq:gbpo_token_obj}. With $n_i=|o_i|$, its unnormalized per-token term and full objective are
\begin{equation}
\label{eq:gbpo_loss_app}
\ell^{\rm token}_{i,t}=-\hat A_i\frac{p_{i,t}}{d_{i,t}},\qquad
\mathcal L_{\rm token}=\frac1G\sum_i\frac1{n_i}\sum_t\ell^{\rm token}_{i,t}.
\end{equation}
Here $d_{i,t}=\max(q_{i,t},\mathrm{sg}(p_{i,t}))$ for $\hat A_i\ge0$ and $d_{i,t}=\max(q_{i,t},1-\mathrm{sg}(p_{i,t}))$ for $\hat A_i<0$. Candidates, advantages, prefixes, reference scores, and denominators are fixed during differentiation.

\paragraph{Step B2: logit gradient.}
Using the softmax derivative for the selected token,
\begin{equation}
\label{eq:gbpo_dr_dz_app}
\frac{\partial(p_{i,t}/d_{i,t})}{\partial z_{i,t,j}}
=\frac{p_{i,t}}{d_{i,t}}\bigl(\mathbb{I}[j=o_{i,t}]-\pi_{i,t,j}\bigr).
\end{equation}
Thus the unnormalized per-token gradient is
\begin{equation}
\label{eq:gbpo_logit_grad_app}
\frac{\partial\ell^{\rm token}_{i,t}}{\partial z_{i,t,j}}
=\hat A_i\frac{p_{i,t}}{d_{i,t}}\bigl(\pi_{i,t,j}-\mathbb{I}[j=o_{i,t}]\bigr).
\end{equation}
Multiplication by $(Gn_i)^{-1}$ gives the full objective's coordinate contribution. For the selected token, this yields Eq.~\eqref{eq:gbpo_token_selected_grad}; $d_{i,t}\ge p_{i,t}$ or $d_{i,t}\ge1-p_{i,t}$ gives the two envelopes in Eq.~\eqref{eq:gbpo_token_bounds}.

\paragraph{Step B3: exact BCE-gradient rescaling.}
Write $p=p_{i,t}$ and $y_i=\mathbb{I}[\hat A_i>0]$. The binary-logit BCE gradient weighted by $|\hat A_i|/(Gn_i)$ is
\begin{equation}
g^{\rm BCE}_{i,t}=\frac{|\hat A_i|}{Gn_i}(p-y_i).
\end{equation}
The selected softmax-logit contribution satisfies
\begin{equation}
\label{eq:gbpo_bce_rescaling_app}
g^{\rm token}_{i,t}=\rho_{i,t}g^{\rm BCE}_{i,t},\qquad
\rho_{i,t}=\begin{cases}
p/d_{i,t},&\hat A_i\ge0,\\
(1-p)/d_{i,t},&\hat A_i<0.
\end{cases}
\end{equation}
For $p\in(0,1)$, both branches have $0<\rho_{i,t}\le1$. For $\hat A_i=0$, both gradients are zero. For a nonzero advantage, the rescaling preserves the BCE-form direction and does not introduce a clipping-induced dead region.

% ============================================================
\subsection{Token-Level Gradient from Sequence-Level GBPO}
\label{app:deriv_gbpo_seq}
Use $P_i,Q_i$ and $D_i$ from Eqs.~\eqref{eq:gbpo_seq_scores}--\eqref{eq:gbpo_norm_neg}, with $n_i=|o_i|$. Since
\begin{equation}
\frac{\partial\log p_{i,t}}{\partial z_{i,t,j}}
=\mathbb{I}[j=o_{i,t}]-\pi_{i,t,j},
\end{equation}
the geometric-mean score has derivative
\begin{equation}
\frac{\partial P_i}{\partial z_{i,t,j}}
=\frac{P_i}{n_i}\bigl(\mathbb{I}[j=o_{i,t}]-\pi_{i,t,j}\bigr).
\end{equation}
With $D_i$ and $\hat A_i$ detached, differentiation of Eq.~\eqref{eq:gbpo_obj} therefore gives
\begin{equation}
\frac{\partial\mathcal L_{\mathrm{GBPO}}}{\partial z_{i,t,j}}
=\frac{\hat A_i}{Gn_i}\frac{P_i}{D_i}
\bigl(\pi_{i,t,j}-\mathbb{I}[j=o_{i,t}]\bigr).
\end{equation}
For the selected coordinate, write $p=p_{i,t}$, so that
\begin{equation}
|g_{i,t}|=\frac{|\hat A_i|}{Gn_i}\frac{P_i}{D_i}(1-p).
\end{equation}
All expressions include the candidate-group mean $1/G$; an additional context-batch mean contributes $1/B$.

\paragraph{Positive branch.}
When $\hat A_i\ge0$, $D_i\ge P_i$, yielding
\begin{equation}
|g_{i,t}|\le\frac{|\hat A_i|}{Gn_i}(1-p)\le\frac{|\hat A_i|}{Gn_i}.
\end{equation}

\paragraph{Negative branch.}
When $\hat A_i<0$, $D_i\ge1-P_i$. Since every token probability is at most one, $P_i\le p^{1/n_i}$. The function $u/(1-u)$ is increasing on $(0,1)$, hence
\begin{align}
\frac{P_i}{1-P_i}(1-p)
&\le\frac{p^{1/n_i}}{1-p^{1/n_i}}(1-p)\\
&=\sum_{j=1}^{n_i}p^{j/n_i},
\end{align}
where the equality uses $1-p=(1-p^{1/n_i})\sum_{j=0}^{n_i-1}p^{j/n_i}$. Consequently,
\begin{equation}
\label{eq:gbpo_seq_neg_bound}
|g_{i,t}|\le\frac{|\hat A_i|}{Gn_i}\sum_{j=1}^{n_i}p^{j/n_i}
\le\frac{|\hat A_i|}{G}p^{1/n_i}\le\frac{|\hat A_i|}{G}.
\end{equation}
These local logit-gradient bounds are independent of $Q_i$. At fixed $n_i$, the negative envelope vanishes as $p\to0$, recovering the token-level envelope at $n_i=1$. The attenuation is quantitatively weaker for longer identifiers; the token-level $O(p)$ envelope must not be substituted for this sequence-level $O(p^{1/n_i})$ envelope.

% ============================================================

\subsection{Active-Region Gradient Coefficients}
\label{app:piecewise_coeffs}
For readability in the main text, we summarize the dependence of logit gradients on the importance ratio.
All methods admit the form
\begin{equation}
\label{eq:piecewise_master}
\frac{\partial \mathcal L}{\partial z_{i,t,j}}
=
\eta\,\kappa_{i,t}\,
\bigl(\pi_{i,t,j}-\mathbb{I}[j=o_{i,t}]\bigr),
\end{equation}
where the normalizer $\eta$ accounts for objective averaging:
$(G|o_i|)^{-1}$ for sequence-normalized GRPO, Dual-Clip PPO, and both GBPO variants,
$G^{-1}$ for sequence-level GSPO,
and $S^{-1}$ for global-token normalized DAPO/GPPO with $S=\sum_{k=1}^G |o_k|$.

\paragraph{GRPO}
\begin{equation}
\label{eq:kappa_grpo_piecewise}
\kappa^{\mathrm{GRPO}}(r,\hat A)=
\begin{cases}
\hat A\,r, & \hat A>0,\ 0<r\le 1+\epsilon,\\
0, & \hat A>0,\ r>1+\epsilon,\\[2pt]
0, & \hat A<0,\ 0<r<1-\epsilon,\\
\hat A\,r, & \hat A<0,\ r\ge 1-\epsilon.
\end{cases}
\end{equation}

\paragraph{DAPO}
\begin{equation}
\label{eq:kappa_dapo_piecewise}
\kappa^{\mathrm{DAPO}}(r,\hat A)=
\begin{cases}
\hat A\,r, & \hat A>0,\ 0<r\le 1+\epsilon_{\mathrm{high}},\\
0, & \hat A>0,\ r>1+\epsilon_{\mathrm{high}},\\[2pt]
0, & \hat A<0,\ 0<r<1-\epsilon_{\mathrm{low}},\\
\hat A\,r, & \hat A<0,\ r\ge 1-\epsilon_{\mathrm{low}}.
\end{cases}
\end{equation}

\paragraph{GSPO}
Let $\tilde r_i$ be defined in \eqref{eq:gspo_si_app}. Then
\begin{equation}
\label{eq:kappa_gspo_piecewise}
\kappa^{\mathrm{GSPO}}(\tilde r, A^b)=
\begin{cases}
A^b\,\dfrac{\tilde r}{|o_i|}, & A^b>0,\ 0<\tilde r\le 1+\epsilon,\\
0, & A^b>0,\ \tilde r>1+\epsilon,\\[6pt]
0, & A^b<0,\ 0<\tilde r<1-\epsilon,\\
A^b\,\dfrac{\tilde r}{|o_i|}, & A^b<0,\ \tilde r\ge 1-\epsilon.
\end{cases}
\end{equation}

\paragraph{GPPO}
Under the stop-gradient clip in \eqref{eq:gppo_rbar_app}, the outer saturation regions
yield constant effective coefficients (rather than zero-gradient truncation):
\begin{equation}
\label{eq:kappa_gppo_piecewise}
\kappa^{\mathrm{GPPO}}(r,\hat A)=
\begin{cases}
\hat A\,r, & \hat A>0,\ 0<r\le 1+\epsilon_{\mathrm{high}},\\
\hat A\,(1+\epsilon_{\mathrm{high}}), & \hat A>0,\ r>1+\epsilon_{\mathrm{high}},\\[2pt]
\hat A\,(1-\epsilon_{\mathrm{low}}), & \hat A<0,\ 0<r<1-\epsilon_{\mathrm{low}},\\
\hat A\,r, & \hat A<0,\ r\ge 1-\epsilon_{\mathrm{low}}.
\end{cases}
\end{equation}

\paragraph{Dual-Clip PPO}
With floor parameter $c$, the negative-advantage branch truncates gradients once $r$ exceeds $c$:
\begin{equation}
\label{eq:kappa_dual_piecewise}
\kappa^{\mathrm{DUAL}}(r,\hat A)=
\begin{cases}
\hat A\,r, & \hat A>0,\ 0<r\le 1+\epsilon,\\
0, & \hat A>0,\ r>1+\epsilon,\\[2pt]
0, & \hat A<0,\ 0<r<1-\epsilon,\\
\hat A\,r, & \hat A<0,\ 1-\epsilon\le r<c,\\
0, & \hat A<0,\ r\ge c.
\end{cases}
\end{equation}

\paragraph{GBPO}
Both variants use continuous stop-gradient normalization instead of hard clipping. With the common normalizer $\eta=(Gn_i)^{-1}$, their coefficient maps are
\begin{equation}
\label{eq:kappa_gbpo_piecewise}
\kappa^{\rm token}_{i,t}=\hat A_i\frac{p_{i,t}}{d_{i,t}},\qquad
\kappa^{\rm seq}_{i,t}=\hat A_i\frac{P_i}{D_i}.
\end{equation}
The token denominator is defined in Section~\ref{subsec:gbpo}; the sequence denominator is given in Eqs.~\eqref{eq:gbpo_norm_pos}--\eqref{eq:gbpo_norm_neg}. For nonzero advantages and interior probabilities, both coefficients remain nonzero. Their selected-coordinate envelopes differ as stated in Eqs.~\eqref{eq:gbpo_token_bounds} and \eqref{eq:gbpo_bound_neg}.

\subsection{Qualitative Analysis of Diagnostic Criteria}
\label{app:qualitative_analysis}
In this section, we provide a detailed justification for the qualitative assessments in Table~\ref{tab:rl_objective_comparison}. The analysis centers on the selected-token logit gradient magnitude $|g| = |\eta \cdot \kappa \cdot (1-p)|$ derived in Sec.~\ref{app:piecewise_coeffs}, where $\eta$ is the normalization factor. Below, $p=\pi_{i,t}$ denotes the selected-token probability, and $r$ denotes the ratio used by each baseline ($r_{i,t}$ for token-level methods and $\tilde r_i$ for GSPO). For asymmetric clipping, $1-\epsilon$ and $1+\epsilon$ refer to the corresponding method-specific lower and upper thresholds.

\paragraph{Full Sample Utilization}
\begin{itemize}
    \item \textbf{GRPO/DAPO/GSPO/Dual-Clip PPO (No):} These methods use hard ratio-based clipping (e.g., $\operatorname{clip}(r, 1-\epsilon, 1+\epsilon)$). Once $r$ enters the saturation region (e.g., $r > 1+\epsilon$ for positive advantages), the surrogate becomes flat w.r.t.\ $r$, leading to $\kappa=0$ (dead gradients). This discards information from sampled tokens, or entire sequences in GSPO, that have moved beyond the clip threshold.
    \item \textbf{GPPO (Yes):} GPPO replaces the zero-gradient region with a stop-gradient construction (typically $\epsilon \cdot r$). This ensures that $\kappa$ remains non-zero globally, allowing all samples to contribute to the update.
    \item \textbf{GBPO, both variants (Yes):} The token and sequence coefficients in Eq.~\eqref{eq:kappa_gbpo_piecewise} are nonzero for nonzero advantages and interior probabilities. Neither construction introduces a clipping-induced dead region. Zero advantages and boundary probabilities are excluded from this sample-activity claim.
\end{itemize}

\paragraph{Positive/Negative Gradient Boundedness}
\begin{itemize}
    \item \textbf{Positive Gradient (All - Yes):} For $\hat A>0$, the clipping-based coefficients remain bounded in their active regions. Both GBPO variants have $|\kappa|\le|\hat A|$, because $d_{i,t}\ge p_{i,t}$ for the token form and $D_i\ge P_i$ for the sequence form. Their full selected-coordinate contributions are bounded by $|\hat A_i|(1-p)/(Gn_i)$.
    \item \textbf{Negative Gradient (Dual-Clip PPO, both GBPO variants - Yes; Others - No):} In standard ratio-based variants, ratio-proportional negative updates can become unbounded as the reference probability approaches zero. Dual-Clip PPO truncates extreme negative updates. Token-level GBPO satisfies $|g^{\rm token}_{i,t}|\le|\hat A_i|p/(Gn_i)$, while sequence-level GBPO satisfies Eq.~\eqref{eq:gbpo_seq_neg_bound}, in particular $|g_{i,t}|\le|\hat A_i|/G$. Both GBPO bounds are independent of reference scores, although their confidence attenuation differs.
\end{itemize}

\paragraph{Confidence Weighting (Positive $\hat A > 0$)}
\begin{itemize}
    \item \textbf{GRPO/DAPO/GSPO/Dual-Clip PPO (Partial):} These methods suffer from dead-gradient regions ($\kappa=0$) when the ratio exceeds the clip threshold. In the active region ($r < 1+\epsilon$), $|g|$ is bounded by a method-specific constant times $(1+\epsilon)(1-p)$, which vanishes as $p \to 1$. However, the vanishing mechanism is partly driven by ratio-triggered hard cutoffs rather than purely by absolute confidence, justifying the ``Partial'' assessment.
    \item \textbf{GPPO (Yes):} GPPO eliminates dead-gradient regions via a stop-gradient construction, ensuring $\kappa \neq 0$ globally. In both the active and saturation regions, $|g|$ is upper-bounded by a term proportional to $(1-p)$, and this decay is purely confidence-driven without any ratio-triggered cutoffs.
    \item \textbf{GBPO, both variants (Yes):} The selected-coordinate envelope is $|\hat A_i|(1-p)/(Gn_i)$ for both forms. It vanishes as $p\to1$ without a ratio-triggered cutoff; see Eqs.~\eqref{eq:gbpo_token_bounds} and \eqref{eq:gbpo_bound_pos}.
\end{itemize}

\paragraph{Confidence Weighting (Negative $\hat A < 0$)}
\begin{itemize}
    \item \textbf{Hard-Clipping Variants (GRPO/DAPO/GSPO) (No):} For negative advantages, these methods are active only when $r \ge 1-\epsilon$. In the active region, the selected-token logit-gradient magnitude behaves as $|g| \propto r(1-p)$, with the method-specific advantage and normalization factors retained. As $p \to 0$ (for already-suppressed negatives), $(1-p)$ approaches one, so small $p$ alone does not guarantee a small update independently of the reference policy; the update scale still depends on $r$. Once $r < 1-\epsilon$, gradients truncate to zero, yielding a ratio-triggered hard cutoff.
    \item \textbf{GPPO (No):} Similarly, in its saturation region, the coefficient $\kappa$ is constant, so $|g| \propto (1-p)$ approaches its \emph{maximum} as $p \to 0$, providing the strongest updates for cases that are already well-handled.
    \item \textbf{GBPO, token form (Yes):} The envelope $|g^{\rm token}_{i,t}|\le|\hat A_i|p/(Gn_i)$ vanishes linearly as $p\to0$, independently of the reference probability; see Eq.~\eqref{eq:gbpo_token_bounds}.
    \item \textbf{GBPO, sequence form (Yes, length-dependent):} The envelope is $|g_{i,t}|\le|\hat A_i|\sum_{j=1}^{n_i}p^{j/n_i}/(Gn_i)\le|\hat A_i|p^{1/n_i}/G$. It vanishes at fixed length, but more slowly than the token-level envelope when $n_i>1$; see Eq.~\eqref{eq:gbpo_seq_neg_bound}.
\end{itemize}

\endgroup

% ============================
% Appendix: other datasets
% ============================
\section{Training Metrics on Other Datasets}
\label{app:other_dataset_metrics}

In the main text, we present training dynamics and stability diagnostics on Sports for clarity and space.
To verify that our findings are not dataset-specific, we additionally report the same set of diagnostics on Beauty and Toys in this appendix.
All plots are generated under the \emph{controlled low-LR} regime (Section~\ref{subsec:stability_dynamics}), using the same backbone, constrained SID decoding, frozen DIN reward, and candidate generation pipeline as in the main experiments.
Specifically, we include: (i) validation HitRate@10 curves (training dynamics), (ii) entropy evolution, (iii) discarded-gradient ratio for clipping/truncation-based methods, and (iv) gradient-norm ratio between negative- and positive-advantage updates.
The definitions of entropy, discarded-gradient ratio, and gradient-norm logging follow exactly those in Section~\ref{subsec:stability_dynamics}.

\subsection{Beauty}
\label{app:beauty_metrics}

As shown in Figures~\ref{fig:beauty_dynamics}--\ref{fig:beauty_grad_ratio}, the training dynamics, entropy evolution, sample utilization, and gradient stability on Beauty are qualitatively consistent with the observations on Sports in the main text. Specifically, GBPO maintains superior stability, remains fully gradient-active, and effectively suppresses negative-side dominance through its confidence-aware design.

\begin{center}
  \includegraphics[width=0.75\linewidth]{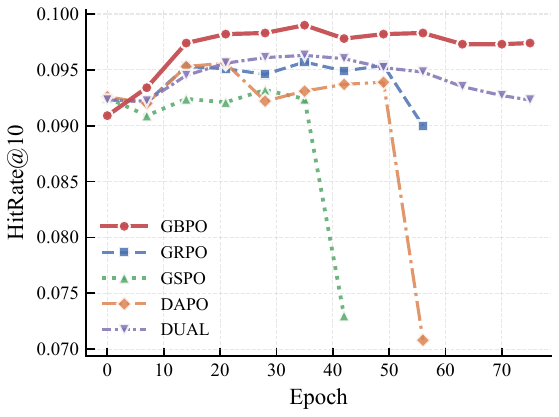}  \captionof{figure}{Comparison of training dynamics across RL objectives in Amazon Beauty.}
  \Description{Validation HitRate@10 over training steps for various RL objectives on the Beauty dataset.}
  \label{fig:beauty_dynamics}
\end{center}

\begin{center}
\includegraphics[width=0.75\linewidth]{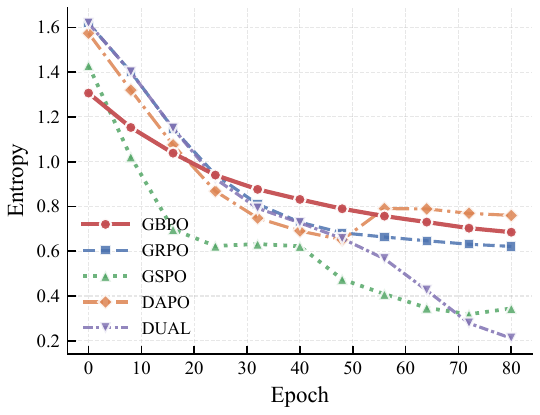}
  \captionof{figure}{{Comparison of entropy evolution across RL methods in Amazon Beauty.} }
  \Description{Entropy evolution over training epochs for different reinforcement learning methods on the Beauty dataset.}
  \label{fig:beauty_entropy}
\end{center}

\begin{center}
  \includegraphics[width=0.75\linewidth]{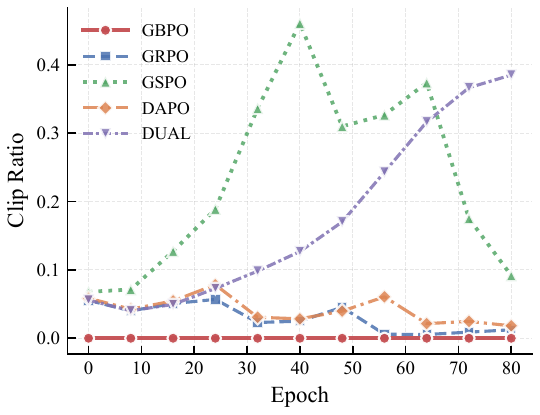}
  \captionof{figure}{{Proportion of samples with discarded gradients across RL methods in Amazon Beauty.} }
  \Description{Plot showing the proportion of samples with discarded gradients across training for various RL methods on the Beauty dataset.}
  \label{fig:beauty_clip}
\end{center}

\begin{center}
  \includegraphics[width=0.75\linewidth]{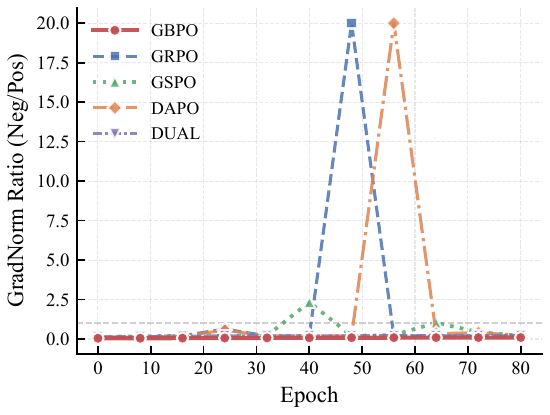}
  \captionof{figure}{{Gradient-Norm Ratio (Negative vs. Positive Advantage) across RL Methods in Amazon Beauty}.}
  \Description{Plot of the gradient-norm ratio between negative and positive advantage updates on the Beauty dataset.}
  \label{fig:beauty_grad_ratio}
\end{center}

\subsection{Toys}
\label{app:toys_metrics}

Similarly, on the Toys dataset (Figures~\ref{fig:toys_dynamics}--\ref{fig:toys_grad_ratio}), GBPO exhibits the same characteristic stability and optimization efficiency. The results further validate that the three diagnostic criteria---full sample utilization, gradient boundedness, and confidence weighting---consistently explain the superior performance of GBPO across diverse recommendation benchmarks.

\begin{center}
  \includegraphics[width=0.75\linewidth]{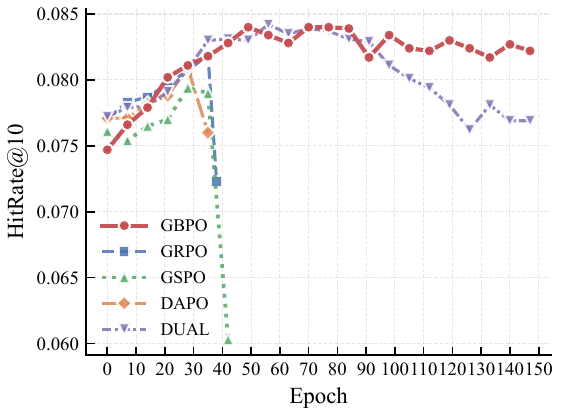}
  \captionof{figure}{Comparison of training dynamics across RL objectives in Amazon Toys.}
  \Description{Validation HitRate@10 over training steps for various RL objectives on the Toys dataset.}
  \label{fig:toys_dynamics}
\end{center}

\begin{center}
  \includegraphics[width=0.75\linewidth]{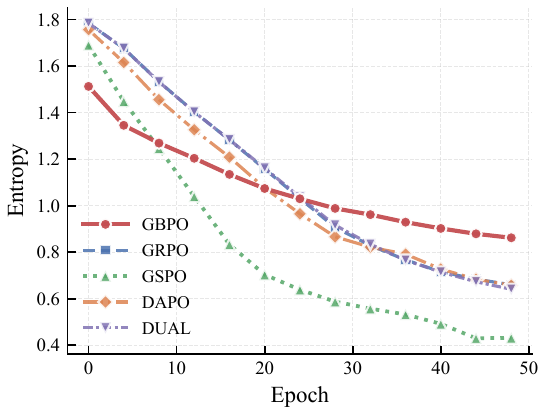}
  \captionof{figure}{{Comparison of entropy evolution across RL methods in Amazon Toys.} }
  \Description{Entropy evolution over training epochs for different reinforcement learning methods on the Toys dataset.}
  \label{fig:toys_entropy}
\end{center}

\begin{center}
  \includegraphics[width=0.75\linewidth]{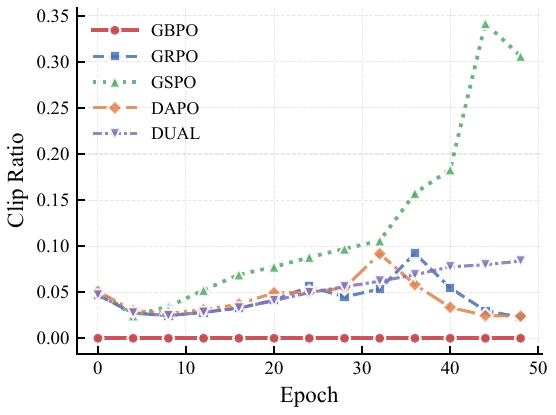}
  \captionof{figure}{{Proportion of samples with discarded gradients across RL methods in Amazon Toys.} }
  \Description{Plot showing the proportion of samples with discarded gradients across training for various RL methods on the Toys dataset.}
  \label{fig:toys_clip}
\end{center}

\begin{center}
  \includegraphics[width=0.75\linewidth]{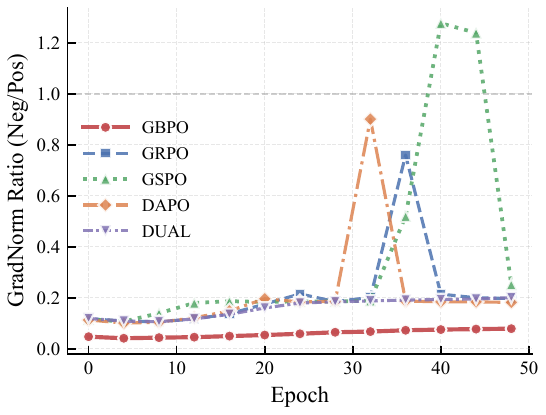}
  \captionof{figure}{{Gradient-Norm Ratio (Negative vs. Positive Advantage) across RL Methods in Amazon Toys.}}
  \Description{Plot of the gradient-norm ratio between negative and positive advantage updates on the Toys dataset.}
  \label{fig:toys_grad_ratio}
\end{center}

\paragraph{Takeaway.}
Across Beauty/Toys/Sports, the same three diagnostics (dead gradients, negative dominance, and confidence-aware weighting) consistently explain the observed stability differences, providing cross-dataset evidence for the design choices of GBPO.

\section{Extended Reward Sensitivity Results}
\label{app:reward_sensitivity_results}

\subsection{Continuous DIN Rewards}
\label{app:extended_reward_mapping}
Table~\ref{tab:continuous_din_results} reports the four recommendation metrics and event markers for all six objectives on Beauty, Toys, and Sports. The binary-DIN reward mapping is replaced with continuous DIN sigmoid scores, retaining the TIGER student and candidate interface.

\begin{table}[t]
\centering
\small
\setlength{\tabcolsep}{5pt}
\renewcommand{\arraystretch}{1.15}
\caption{Single-run results with continuous DIN sigmoid rewards on TIGER. Best values within each dataset are bold.}
\label{tab:continuous_din_results}
\begin{tabular}{llrrrr}
\toprule
\textbf{Dataset} & \textbf{Method} & \textbf{H@5} & \textbf{H@10} & \textbf{NDCG@5} & \textbf{NDCG@10} \\
\midrule
\textbf{Beauty} & GRPO$^{\dagger}$ & 0.0413 & 0.0647 & 0.0259 & 0.0334\\
 & DAPO$^{\dagger}$ & 0.0415 & 0.0640 & 0.0267 & 0.0339\\
 & GSPO$^{\dagger}$ & \textbf{0.0427} & 0.0662 & 0.0278 & 0.0353\\
 & GPPO$^{\dagger}$ & 0.0372 & 0.0624 & 0.0235 & 0.0316\\
 & Dual-Clip$^{\dagger}$ & 0.0422 & 0.0667 & \textbf{0.0280} & \textbf{0.0359}\\
 & \textbf{GBPO} & 0.0405 & \textbf{0.0687} & 0.0243 & 0.0333\\
\midrule
\textbf{Toys} & GRPO$^{\dagger}$ & 0.0312 & 0.0521 & 0.0198 & 0.0265\\
 & DAPO$^{\dagger}$ & 0.0327 & 0.0535 & 0.0202 & 0.0269\\
 & GSPO$^{\dagger}$ & 0.0326 & 0.0551 & 0.0203 & 0.0276\\
 & GPPO$^{\dagger}$ & 0.0321 & 0.0543 & 0.0200 & 0.0272\\
 & Dual-Clip$^{\dagger}$ & 0.0322 & 0.0531 & 0.0197 & 0.0264\\
 & \textbf{GBPO} & \textbf{0.0421} & \textbf{0.0698} & \textbf{0.0256} & \textbf{0.0346}\\
\midrule
\textbf{Sports} & GRPO$^{\dagger}$ & 0.0226 & 0.0379 & 0.0143 & 0.0192\\
 & DAPO$^{\dagger}$ & 0.0233 & 0.0380 & 0.0144 & 0.0191\\
 & GSPO$^{\dagger}$ & 0.0242 & 0.0405 & 0.0152 & 0.0204\\
 & GPPO$^{\dagger\ddagger}$ & 0.0226 & 0.0380 & 0.0144 & 0.0193\\
 & Dual-Clip$^{\dagger\ddagger}$ & 0.0219 & 0.0367 & 0.0137 & 0.0184\\
 & \textbf{GBPO} & \textbf{0.0282} & \textbf{0.0459} & \textbf{0.0180} & \textbf{0.0237}\\
\bottomrule
\end{tabular}
\end{table}

\subsection{Alternative Ranking Teacher}
\label{app:extended_teacher}
Table~\ref{tab:sasrec_teacher_results} reports the corresponding single-run sensitivity comparison using a separately trained SASRec-style teacher with binary rewards and the same TIGER student and candidate interface.

\begin{table}[t]
\centering
\small
\setlength{\tabcolsep}{5pt}
\renewcommand{\arraystretch}{1.15}
\caption{Single-run results with a SASRec-style teacher and binary rewards on TIGER. Best values within each dataset are bold.}
\label{tab:sasrec_teacher_results}
\begin{tabular}{llrrrr}
\toprule
\textbf{Dataset} & \textbf{Method} & \textbf{H@5} & \textbf{H@10} & \textbf{NDCG@5} & \textbf{NDCG@10} \\
\midrule
\textbf{Beauty} & GRPO$^{\dagger}$ & 0.0470 & 0.0753 & 0.0310 & 0.0401\\
 & DAPO$^{\dagger}$ & 0.0474 & 0.0744 & 0.0313 & 0.0400\\
 & GSPO$^{\dagger}$ & \textbf{0.0523} & 0.0788 & \textbf{0.0340} & 0.0426\\
 & GPPO$^{\dagger}$ & 0.0478 & 0.0745 & 0.0297 & 0.0383\\
 & Dual-Clip$^{\dagger}$ & 0.0516 & 0.0793 & 0.0339 & \textbf{0.0428}\\
 & \textbf{GBPO} & 0.0522 & \textbf{0.0796} & 0.0327 & 0.0415\\
\midrule
\textbf{Toys} & GRPO$^{\dagger}$ & 0.0416 & 0.0638 & 0.0260 & 0.0332\\
 & DAPO$^{\dagger}$ & 0.0441 & 0.0678 & 0.0279 & 0.0355\\
 & GSPO$^{\dagger}$ & 0.0580 & 0.0851 & 0.0360 & 0.0447\\
 & GPPO$^{\dagger}$ & 0.0560 & 0.0831 & 0.0355 & 0.0442\\
 & Dual-Clip$^{\dagger}$ & 0.0534 & 0.0792 & 0.0334 & 0.0417\\
 & \textbf{GBPO} & \textbf{0.0597} & \textbf{0.0869} & \textbf{0.0380} & \textbf{0.0468}\\
\midrule
\textbf{Sports} & GRPO$^{\dagger}$ & 0.0288 & 0.0443 & 0.0181 & 0.0232\\
 & DAPO$^{\dagger}$ & 0.0279 & 0.0448 & 0.0180 & 0.0235\\
 & GSPO$^{\dagger}$ & 0.0308 & 0.0480 & 0.0197 & 0.0253\\
 & GPPO$^{\dagger}$ & 0.0265 & 0.0437 & 0.0168 & 0.0224\\
 & Dual-Clip$^{\dagger\ddagger}$ & 0.0305 & 0.0464 & 0.0194 & 0.0245\\
 & \textbf{GBPO} & \textbf{0.0310} & \textbf{0.0485} & \textbf{0.0202} & \textbf{0.0258}\\
\bottomrule
\end{tabular}
\end{table}

\clearpage
\section{Amazon Replications on a New Server}
\label{app:amazon_replications}
We repeat the default binary-DIN RL comparisons on Beauty, Toys, and Sports with TIGER and HSTU on a new server. Seeds 42--44 vary the RL stage, with a fixed SFT initialization and DIN teacher within each backbone--dataset setting. The original single-run Amazon results remain in the main text; the following tables report the new-server experiments separately with their corresponding SFT references.

\subsection{TIGER}
\label{app:tiger_replication}
\begin{table}[t]
\centering
\small
\setlength{\tabcolsep}{3pt}
\renewcommand{\arraystretch}{1.15}
\caption{TIGER results under default binary-DIN rewards. RL scores are mean $\pm$ sample standard deviation over seeds 42--44. SFT is the fixed initialization reference. The highest RL means within each dataset are bold.}
\label{tab:tiger_three_seed_replication}
\begin{tabular}{llccccr}
\toprule
\textbf{Dataset} & \textbf{Method} & \textbf{H@5} & \textbf{H@10} & \textbf{NDCG@5} & \textbf{NDCG@10} & \textbf{D/N} \\
\midrule
\textbf{Beauty} & SFT & 0.0438 & 0.0682 & 0.0295 & 0.0374 &  \\
 & GRPO & $0.0478\pm0.0014$ & $0.0734\pm0.0006$ & $0.0315\pm0.0010$ & $0.0397\pm0.0007$ & 3/0 \\
 & DAPO & $0.0475\pm0.0003$ & $0.0732\pm0.0015$ & $0.0316\pm0.0003$ & $0.0398\pm0.0006$ & 3/0 \\
 & GSPO & $0.0492\pm0.0007$ & $0.0761\pm0.0016$ & $0.0317\pm0.0004$ & $0.0404\pm0.0003$ & 3/0 \\
 & GPPO & $0.0462\pm0.0003$ & $0.0731\pm0.0008$ & $0.0292\pm0.0003$ & $0.0379\pm0.0004$ & 3/0 \\
 & Dual-Clip & $0.0487\pm0.0010$ & $0.0736\pm0.0016$ & $0.0324\pm0.0003$ & $0.0404\pm0.0005$ & 3/0 \\
 & \textbf{GBPO} & $\mathbf{0.0519}\pm\mathbf{0.0016}$ & $\mathbf{0.0775}\pm\mathbf{0.0011}$ & $\mathbf{0.0329}\pm\mathbf{0.0010}$ & $\mathbf{0.0412}\pm\mathbf{0.0008}$ & 0/0 \\
\midrule
\textbf{Toys} & SFT & 0.0361 & 0.0583 & 0.0228 & 0.0300 &  \\
 & GRPO & $0.0389\pm0.0006$ & $0.0608\pm0.0016$ & $0.0248\pm0.0004$ & $0.0319\pm0.0004$ & 3/0 \\
 & DAPO & $0.0407\pm0.0006$ & $0.0638\pm0.0014$ & $0.0256\pm0.0004$ & $0.0330\pm0.0008$ & 3/0 \\
 & GSPO & $0.0463\pm0.0017$ & $0.0715\pm0.0014$ & $0.0295\pm0.0007$ & $0.0376\pm0.0006$ & 3/0 \\
 & GPPO & $0.0446\pm0.0018$ & $0.0688\pm0.0006$ & $0.0281\pm0.0013$ & $0.0359\pm0.0008$ & 3/0 \\
 & Dual-Clip & $0.0432\pm0.0005$ & $0.0664\pm0.0013$ & $0.0275\pm0.0010$ & $0.0350\pm0.0004$ & 3/0 \\
 & \textbf{GBPO} & $\mathbf{0.0536}\pm\mathbf{0.0005}$ & $\mathbf{0.0771}\pm\mathbf{0.0011}$ & $\mathbf{0.0349}\pm\mathbf{0.0006}$ & $\mathbf{0.0426}\pm\mathbf{0.0007}$ & 0/0 \\
\midrule
\textbf{Sports} & SFT & 0.0230 & 0.0391 & 0.0146 & 0.0197 &  \\
 & GRPO & $0.0271\pm0.0003$ & $0.0417\pm0.0001$ & $0.0170\pm0.0003$ & $0.0217\pm0.0003$ & 3/0 \\
 & DAPO & $0.0267\pm0.0003$ & $0.0418\pm0.0005$ & $0.0170\pm0.0003$ & $0.0218\pm0.0004$ & 3/0 \\
 & GSPO & $0.0279\pm0.0005$ & $0.0444\pm0.0007$ & $0.0174\pm0.0004$ & $0.0228\pm0.0003$ & 3/0 \\
 & GPPO & $0.0272\pm0.0008$ & $0.0442\pm0.0001$ & $0.0170\pm0.0002$ & $0.0225\pm0.0001$ & 3/0 \\
 & Dual-Clip & $0.0275\pm0.0004$ & $0.0437\pm0.0010$ & $0.0173\pm0.0003$ & $0.0225\pm0.0005$ & 3/3 \\
 & \textbf{GBPO} & $\mathbf{0.0300}\pm\mathbf{0.0003}$ & $\mathbf{0.0478}\pm\mathbf{0.0003}$ & $\mathbf{0.0188}\pm\mathbf{0.0003}$ & $\mathbf{0.0245}\pm\mathbf{0.0002}$ & 0/0 \\
\bottomrule
\end{tabular}
\end{table}

\subsection{HSTU}
\label{app:hstu_replication}
\begin{table}[t]
\centering
\small
\setlength{\tabcolsep}{3pt}
\renewcommand{\arraystretch}{1.15}
\caption{HSTU results under default binary-DIN rewards. RL scores are mean $\pm$ sample standard deviation over seeds 42--44. SFT is the fixed initialization reference. The highest RL means within each dataset are bold.}
\label{tab:hstu_three_seed_replication}
\begin{tabular}{llccccr}
\toprule
\textbf{Dataset} & \textbf{Method} & \textbf{H@5} & \textbf{H@10} & \textbf{NDCG@5} & \textbf{NDCG@10} & \textbf{D/N} \\
\midrule
\textbf{Beauty} & SFT & 0.0443 & 0.0617 & 0.0316 & 0.0372 &  \\
 & GRPO & $0.0483\pm0.0012$ & $0.0686\pm0.0006$ & $0.0331\pm0.0008$ & $0.0396\pm0.0006$ & 3/0 \\
 & DAPO & $0.0496\pm0.0009$ & $0.0704\pm0.0010$ & $0.0339\pm0.0004$ & $0.0406\pm0.0005$ & 3/0 \\
 & GSPO & $0.0483\pm0.0012$ & $0.0686\pm0.0006$ & $0.0331\pm0.0008$ & $0.0396\pm0.0006$ & 3/0 \\
 & GPPO & $0.0467\pm0.0002$ & $0.0724\pm0.0006$ & $0.0307\pm0.0003$ & $0.0390\pm0.0003$ & 0/0 \\
 & Dual-Clip & $\mathbf{0.0543}\pm\mathbf{0.0018}$ & $\mathbf{0.0784}\pm\mathbf{0.0022}$ & $\mathbf{0.0369}\pm\mathbf{0.0008}$ & $\mathbf{0.0447}\pm\mathbf{0.0008}$ & 1/2 \\
 & \textbf{GBPO} & $0.0495\pm0.0015$ & $0.0738\pm0.0009$ & $0.0324\pm0.0009$ & $0.0402\pm0.0006$ & 0/0 \\
\midrule
\textbf{Toys} & SFT & 0.0414 & 0.0539 & 0.0304 & 0.0345 &  \\
 & GRPO & $0.0474\pm0.0007$ & $0.0653\pm0.0010$ & $0.0331\pm0.0008$ & $0.0388\pm0.0006$ & 3/0 \\
 & DAPO & $0.0456\pm0.0039$ & $0.0624\pm0.0044$ & $0.0319\pm0.0027$ & $0.0373\pm0.0029$ & 3/0 \\
 & GSPO & $0.0474\pm0.0007$ & $0.0653\pm0.0010$ & $0.0331\pm0.0008$ & $0.0388\pm0.0006$ & 3/0 \\
 & GPPO & $0.0469\pm0.0022$ & $0.0706\pm0.0013$ & $0.0310\pm0.0015$ & $0.0386\pm0.0011$ & 0/0 \\
 & Dual-Clip & $\mathbf{0.0544}\pm\mathbf{0.0009}$ & $\mathbf{0.0760}\pm\mathbf{0.0017}$ & $\mathbf{0.0374}\pm\mathbf{0.0002}$ & $\mathbf{0.0443}\pm\mathbf{0.0005}$ & 0/3 \\
 & \textbf{GBPO} & $0.0471\pm0.0014$ & $0.0705\pm0.0015$ & $0.0313\pm0.0011$ & $0.0389\pm0.0012$ & 0/0 \\
\midrule
\textbf{Sports} & SFT & 0.0200 & 0.0301 & 0.0134 & 0.0167 &  \\
 & GRPO & $0.0290\pm0.0004$ & $0.0447\pm0.0004$ & $0.0188\pm0.0003$ & $0.0239\pm0.0002$ & 3/0 \\
 & DAPO & $0.0292\pm0.0002$ & $0.0447\pm0.0003$ & $\mathbf{0.0189}\pm\mathbf{0.0000}$ & $0.0239\pm0.0001$ & 2/1 \\
 & GSPO & $0.0290\pm0.0004$ & $0.0447\pm0.0004$ & $0.0188\pm0.0003$ & $0.0239\pm0.0002$ & 3/0 \\
 & GPPO & $0.0269\pm0.0006$ & $0.0438\pm0.0005$ & $0.0170\pm0.0005$ & $0.0224\pm0.0004$ & 0/2 \\
 & Dual-Clip & $\mathbf{0.0293}\pm\mathbf{0.0006}$ & $\mathbf{0.0457}\pm\mathbf{0.0003}$ & $\mathbf{0.0189}\pm\mathbf{0.0002}$ & $\mathbf{0.0241}\pm\mathbf{0.0001}$ & 1/2 \\
 & \textbf{GBPO} & $0.0285\pm0.0002$ & $0.0446\pm0.0007$ & $0.0179\pm0.0003$ & $0.0231\pm0.0003$ & 0/0 \\
\bottomrule
\end{tabular}
\end{table}

\section{Statistical Significance Analysis}
\label{app:statistical_significance}
Table~\ref{tab:statistical_significance} reports Holm-adjusted $p$-values computed from the per-seed scores. GBPO significantly outperforms baselines in 28 of 60 TIGER comparisons. On HSTU, only Toys H@10 versus Dual-Clip is significant, favoring Dual-Clip; no Steam comparison reaches significance.

With only three seeds, these tests have limited power and depend on distributional assumptions; single-run comparisons remain descriptive.

\begin{table}[t]
\centering
\footnotesize
\setlength{\tabcolsep}{4pt}
\renewcommand{\arraystretch}{1.08}
\caption{Holm-adjusted $p$-values for GBPO versus five RL baselines. Bold indicates $p<0.05$; $\uparrow/\downarrow$ denotes a higher/lower GBPO mean in the corresponding result table.}
\label{tab:statistical_significance}
\begin{tabular}{lllcccc}
\toprule
\textbf{Backbone} & \textbf{Dataset} & \textbf{Baseline} & \textbf{H@5} & \textbf{H@10} & \textbf{NDCG@5} & \textbf{NDCG@10} \\
\midrule
\textbf{TIGER} & \textbf{Beauty} & GRPO & $0.66654^{\uparrow}$ & $0.16281^{\uparrow}$ & $0.77618^{\uparrow}$ & $0.12631^{\uparrow}$ \\
 &  & DAPO & $0.47382^{\uparrow}$ & $0.06816^{\uparrow}$ & $0.77618^{\uparrow}$ & $0.12631^{\uparrow}$ \\
 &  & GSPO & $0.98035^{\uparrow}$ & $1.00000^{\uparrow}$ & $1.00000^{\uparrow}$ & $1.00000^{\uparrow}$ \\
 &  & GPPO & $0.28041^{\uparrow}$ & $0.28040^{\uparrow}$ & $0.28041^{\uparrow}$ & $0.26830^{\uparrow}$ \\
 &  & Dual-Clip & $0.77618^{\uparrow}$ & $0.56326^{\uparrow}$ & $1.00000^{\uparrow}$ & $0.98035^{\uparrow}$ \\
\midrule
 & \textbf{Toys} & GRPO & $\mathbf{0.01528}^{\uparrow}$ & $\mathbf{0.00692}^{\uparrow}$ & $\mathbf{0.01488}^{\uparrow}$ & $\mathbf{0.00492}^{\uparrow}$ \\
 &  & DAPO & $\mathbf{0.01488}^{\uparrow}$ & $\mathbf{0.04404}^{\uparrow}$ & $\mathbf{0.02640}^{\uparrow}$ & $\mathbf{0.04404}^{\uparrow}$ \\
 &  & GSPO & $\mathbf{0.04404}^{\uparrow}$ & $\mathbf{0.01488}^{\uparrow}$ & $\mathbf{0.01279}^{\uparrow}$ & $\mathbf{0.00360}^{\uparrow}$ \\
 &  & GPPO & $\mathbf{0.04404}^{\uparrow}$ & $\mathbf{0.02640}^{\uparrow}$ & $\mathbf{0.04404}^{\uparrow}$ & $\mathbf{0.02494}^{\uparrow}$ \\
 &  & Dual-Clip & $\mathbf{0.02640}^{\uparrow}$ & $\mathbf{0.00360}^{\uparrow}$ & $\mathbf{0.04404}^{\uparrow}$ & $\mathbf{0.04362}^{\uparrow}$ \\
\midrule
 & \textbf{Sports} & GRPO & $0.06136^{\uparrow}$ & $\mathbf{0.01451}^{\uparrow}$ & $0.08839^{\uparrow}$ & $\mathbf{0.00807}^{\uparrow}$ \\
 &  & DAPO & $0.06136^{\uparrow}$ & $\mathbf{0.04669}^{\uparrow}$ & $0.08839^{\uparrow}$ & $0.08839^{\uparrow}$ \\
 &  & GSPO & $0.08839^{\uparrow}$ & $0.08803^{\uparrow}$ & $0.08839^{\uparrow}$ & $0.08839^{\uparrow}$ \\
 &  & GPPO & $0.08839^{\uparrow}$ & $\mathbf{0.01644}^{\uparrow}$ & $\mathbf{0.01644}^{\uparrow}$ & $\mathbf{0.01644}^{\uparrow}$ \\
 &  & Dual-Clip & $\mathbf{0.00365}^{\uparrow}$ & $0.08839^{\uparrow}$ & $\mathbf{0.00929}^{\uparrow}$ & $0.08839^{\uparrow}$ \\
\midrule
\textbf{HSTU} & \textbf{Beauty} & GRPO & $1.00000^{\uparrow}$ & $0.37386^{\uparrow}$ & $1.00000^{\downarrow}$ & $1.00000^{\uparrow}$ \\
 &  & DAPO & $1.00000^{\downarrow}$ & $1.00000^{\uparrow}$ & $1.00000^{\downarrow}$ & $1.00000^{\downarrow}$ \\
 &  & GSPO & $1.00000^{\uparrow}$ & $0.37386^{\uparrow}$ & $1.00000^{\downarrow}$ & $1.00000^{\uparrow}$ \\
 &  & GPPO & $1.00000^{\uparrow}$ & $1.00000^{\uparrow}$ & $1.00000^{\uparrow}$ & $1.00000^{\uparrow}$ \\
 &  & Dual-Clip & $1.00000^{\downarrow}$ & $1.00000^{\downarrow}$ & $0.68149^{\downarrow}$ & $0.55407^{\downarrow}$ \\
\midrule
 & \textbf{Toys} & GRPO & $1.00000^{\downarrow}$ & $0.98174^{\uparrow}$ & $1.00000^{\downarrow}$ & $1.00000^{\uparrow}$ \\
 &  & DAPO & $1.00000^{\uparrow}$ & $0.72200^{\uparrow}$ & $1.00000^{\downarrow}$ & $1.00000^{\uparrow}$ \\
 &  & GSPO & $1.00000^{\downarrow}$ & $0.98174^{\uparrow}$ & $1.00000^{\downarrow}$ & $1.00000^{\uparrow}$ \\
 &  & GPPO & $1.00000^{\uparrow}$ & $1.00000^{\downarrow}$ & $1.00000^{\uparrow}$ & $1.00000^{\uparrow}$ \\
 &  & Dual-Clip & $0.13714^{\downarrow}$ & $\mathbf{0.01361}^{\downarrow}$ & $0.14194^{\downarrow}$ & $0.10554^{\downarrow}$ \\
\midrule
 & \textbf{Sports} & GRPO & $1.00000^{\downarrow}$ & $1.00000^{\downarrow}$ & $0.08180^{\downarrow}$ & $0.27483^{\downarrow}$ \\
 &  & DAPO & $0.86023^{\downarrow}$ & $1.00000^{\downarrow}$ & $0.34967^{\downarrow}$ & $0.69707^{\downarrow}$ \\
 &  & GSPO & $1.00000^{\downarrow}$ & $1.00000^{\downarrow}$ & $0.08180^{\downarrow}$ & $0.27483^{\downarrow}$ \\
 &  & GPPO & $0.66984^{\uparrow}$ & $0.86336^{\uparrow}$ & $1.00000^{\uparrow}$ & $1.00000^{\uparrow}$ \\
 &  & Dual-Clip & $1.00000^{\downarrow}$ & $1.00000^{\downarrow}$ & $0.69707^{\downarrow}$ & $0.46165^{\downarrow}$ \\
\midrule
\textbf{TIGER} & \textbf{Steam} & GRPO & $1.00000^{\uparrow}$ & $1.00000^{\uparrow}$ & $1.00000^{\uparrow}$ & $1.00000^{\uparrow}$ \\
 &  & DAPO & $1.00000^{\uparrow}$ & $1.00000^{\uparrow}$ & $1.00000^{\uparrow}$ & $1.00000^{\uparrow}$ \\
 &  & GSPO & $1.00000^{\downarrow}$ & $1.00000^{\downarrow}$ & $1.00000^{\downarrow}$ & $1.00000^{\downarrow}$ \\
 &  & GPPO & $1.00000^{\uparrow}$ & $1.00000^{\uparrow}$ & $1.00000^{\uparrow}$ & $1.00000^{\uparrow}$ \\
 &  & Dual-Clip & $1.00000^{\uparrow}$ & $1.00000^{\uparrow}$ & $1.00000^{\uparrow}$ & $1.00000^{\uparrow}$ \\
\bottomrule
\end{tabular}
\end{table}

\end{document}